\documentclass[journal]{IEEEtran}
\ifCLASSINFOpdf
  \usepackage[pdftex]{graphicx}
\else
  \graphicspath{{../eps/}}
\fi
\usepackage{array}
\usepackage[fleqn]{amsmath}
\usepackage{color}
\usepackage{xcolor}
\usepackage{url} 
\usepackage{footnote}
\usepackage{threeparttable}
\usepackage{epstopdf}
\usepackage{booktabs}
\usepackage{subfigure}
\usepackage{verbatim}
\usepackage{graphicx}
\usepackage{multirow}
\usepackage{pifont}
\usepackage{amssymb}
\usepackage[numbers,sort&compress]{natbib}
\usepackage{algorithm}
\usepackage{algorithmicx}
\usepackage{algpseudocode}
\usepackage{setspace}

\begin{document}
\begin{sloppypar}
\title{Perception-oriented Bidirectional Attention Network for Image Super-resolution Quality Assessment}
%$^{\star}$
\author{Yixiao Li,        Xiaoyuan Yang, Guanghui Yue, Jun Fu, Qiuping Jiang,~\IEEEmembership{Senior Member,~IEEE}, \\ Xu Jia, Paul L. Rosin, Hantao Liu, 
        and~Wei Zhou,~\IEEEmembership{Senior Member,~IEEE}% <-this % stops a space
\thanks{This work was in part supported by the National Natural Science Foundation of China under Grant 62371017. Corresponding author: X. Yang and W. Zhou.}
\thanks{Y. Li is with the School of Mathematical Sciences, Beihang University, Beijing 100191, China, and also with the School of Computer Science and Informatics, Cardiff University, Cardiff CF24 4AG, United Kingdom. (e-mail: 18335310648@163.com).}
\thanks{X. Yang is with the School of Mathematical Sciences, Beihang University, Beijing 100191, China (e-mail: xiaoyuanyang@vip.163.com).}
\thanks{G. Yue is with the School of Biomedical Engineering, Shenzhen University, Shenzhen 518060,
China (e-mail: yueguanghui@szu.edu.cn).}
\thanks{J. Fu is with the CAS Key Laboratory of Technology in Geo-Spatial
Information Processing and Application System, University of Science and
Technology of China, Hefei 230027, China (e-mail: fujun@mail.ustc.edu.cn).}
\thanks{Q. Jiang is with the School of Information Science and
Engineering, Ningbo University, Ningbo 315211, China (e-mail: jiangqiuping@nbu.edu.cn).}
\thanks{X. Jia is with the School of Artificial Intelligence, Dalian University of Technology, Dalian 116081, China (email: xjia@dlut.edu.cn).}
\thanks{P.L. Rosin, H. Liu and W. Zhou are with the School of Computer Science and Informatics, Cardiff University, Cardiff CF24 4AG, United Kingdom (email: rosinpl@cardiff.ac.uk; liuh35@cardiff.ac.uk; zhouw26@cardiff.ac.uk).}
}
%\thanks{$^{\star}$ This work was completed during remote internship at Cardiff University.}

\markboth{IEEE Transactions on Image Processing}%
{Shell \MakeLowercase{\textit{et al.}}: Bare Demo of IEEEtran.cls for IEEE Journals}



\maketitle

\begin{abstract}
Many super-resolution (SR) algorithms have been proposed to increase image resolution. However, full-reference (FR) image quality assessment (IQA) metrics for comparing and evaluating different SR algorithms are limited. In this work, we propose the Perception-oriented Bidirectional Attention Network (PBAN) for image SR FR-IQA, which is composed of three modules: an image encoder module, a perception-oriented bidirectional attention (PBA) module, and a quality prediction module. First, we encode the input images for feature representations. Inspired by the characteristics of the human visual system, we then construct the perception-oriented PBA module. Specifically, different from existing attention-based SR IQA methods, we conceive a Bidirectional Attention to bidirectionally construct visual attention to distortion, which is consistent with the generation and evaluation processes of SR images. To further guide the quality assessment towards the perception of distorted information, we propose Grouped Multi-scale Deformable Convolution, enabling the proposed method to adaptively perceive distortion. Moreover, we design Sub-information Excitation Convolution to direct visual perception to both sub-pixel and sub-channel attention. Finally, the quality prediction module is exploited to integrate quality-aware features and regress quality scores. Extensive experiments demonstrate that our proposed PBAN outperforms state-of-the-art quality assessment methods.
\end{abstract}

% \fi
\IEEEpeerreviewmaketitle

\section{Introduction}
\IEEEPARstart{I}{mage} super-resolution (SR) focuses on reconstructing high-frequency details to generate higher-resolution (HR) images that exhibit more refined structures and textures than their low-resolution (LR) counterparts. The latest advent of deep learning has significantly propelled this field, giving rise to a wide array of SR techniques, including convolutional neural network (CNN)-based approaches~\cite{peng2024efficient,lee2025emulating}, generative adversarial network (GAN)-based frameworks~\cite{zhang2019ranksrgan,li2024sed}, Transformer-based strategies~\cite{wang2024camixersr,park2025efficient}, and graph neural network (GNN)-based algorithms~\cite{tian2024image,AttenSR}. Despite notable advances, SR remains a fundamentally ill-posed problem—where one LR image may correspond to multiple plausible HR outputs—making it inherently difficult to guarantee both accuracy and consistency in restoration. To address this, recent research has increasingly incorporated generative priors, particularly through GAN- and diffusion-based models~\cite{zhang2019ranksrgan,li2024sed,yang2025diffusion,li2025difiisr,moser2025dynamic}. While these methods have shown promising results, striking an effective balance between perceptual realism and reconstruction fidelity remains challenging. GAN-based approaches often yield high-fidelity metrics but may fail to capture vivid textures due to their unstable adversarial training and tendency toward over-optimization~\cite{wu2024seesr}. While diffusion models can produce detailed textures by leveraging powerful generative priors, their reliance on stochastic noise sampling and mismatches between prior and LR distributions can undermine pixel-level accuracy~\cite{yu2024scaling}. Therefore, reliably evaluating the perceptual quality of super-resolution images is of great significance for advancing further optimization of the super-resolution approach.

Generally, image quality assessment (IQA) methods are categorized into full-reference (FR), reduced-reference (RR), and blind/no-reference (NR) based on the availability of reference images. 
When the corresponding original image is available, FR-IQA can directly compare the original and distorted images. The RR-IQA uses partial information from the reference image, while the NR-IQA estimates image quality without any reference information. These methodologies aim to obtain quality predictions that closely align with subjective ratings. Among them, the peak signal-to-noise ratio (PSNR) or mean square error (MSE) is the earliest metric, which computes the pixel-level differences without consideration of the human visual system (HVS) characteristics. Afterward, the structural similarity (SSIM) index~\cite{ssim} was developed, and takes into account the brightness, contrast, and structure perceptual elements of images, and has given rise to many variants~\cite{cwssim,msssim}. However, these methods still lack exploration of high-level semantic information. Recently, deep learning-based common IQA methods~\cite{bosse, DBCNN,qiup2, metaiqa, qiup3,hyperiqa, LIQA} have made significant progress by leveraging semantic features to assess image quality. However, these methods are primarily designed and trained on databases containing common degradation-based distortions, such as blur, noise, and compression artifacts. These distortions typically result in loss of image information and perceptual degradation. In contrast, SR methods introduce a different class of distortions—enhancement-induced artifacts—such as over-sharpened edges, hallucinated or false textures, and reconstruction artifacts. Unlike degradations that remove or obscure image content, these distortions may introduce artificial structures or exaggerated details that can mislead traditional IQA models~\cite{gu2020image}. The existing IQA methods, biased toward identifying degradative patterns, lack the capacity to effectively recognize and penalize these subtle yet perceptually significant errors~\cite{9937738}. Therefore, evaluating the quality of SR images demands specifically designed IQA models that can capture the nuanced characteristics of enhancement-induced distortions.

Except for common IQA, some works have emerged for developing SR IQA methods. The recent SR FR-IQA methods~\cite{sis,sfsn,srif} predominantly rely on hand-crafted features. Although these methods have been proposed to tackle SR image artifacts like jagged edges, blurring, or non-existent artifacts in the original scene, their inability to analyze high-level semantic features curtails their performance. Meanwhile, there have been many attempts to apply deep learning to SR IQA, including CNNs~\cite{deepsqr, HLSRIQA,disq}, knowledge distillation~\cite{eksriqa}, attention mechanisms~\cite{jcsan,tadsrnet}, etc., but these are designed for either NR or RR scenarios. Therefore, there remains a significant gap in exploring deep-learning-based approaches for SR FR-IQA.

Inspired by the process of subjective quality assessment, we propose assessing the artifacts introduced by enhancement (i.e., image SR) from two perspectives: the “generate distortion” and “evaluate distortion” processes. Initially, acquiring SR images from original reference images or real-world scenarios introduces unique enhanced distortions. Then, subjective evaluations by individuals assess such distortion by comparing SR images against the original reference images, or implicitly with hallucinated scenes in mind.  Inspired by the dual aspects of SR image distortion—“generation" and “evaluation"—we propose the \textbf{Bidirectional Attention (Bi-Atten)}, which aligns SR images with their corresponding references in two directions (HR→SR and SR→HR). Unlike vanilla spatial or channel attention mechanisms commonly used in SR NR-IQA methods, it avoids pooling operations, but uses the multiplication of feature matrices from both SR and reference sources to align the two feature spaces.

Besides, we obtain the effective perception of distortion from the following two aspects, and conceive the overall Perception-oriented Bidirectional Attention (PBA). On the one hand, we upgrade deformable convolution~\cite{dai_deformable_2017,zhu_deformable_2019} to adaptively offer the perception of distorted areas. Deformable convolutions can effectively perform dense spatial prediction due to their adaptive sampling mechanism, which aligns with the spatial frequency sensitivity~\cite{spatialofHVS} of the HVS. Therefore, it has been a viable candidate for preliminary integration within existing IQA methods~\cite{radn,dqm-iqa}. Nonetheless, an inherent challenge persists in striking an optimal balance between sampling efficiency and computational complexity within these deformable convolutions. Under this consideration, we devise a novel architecture named \textbf{Grouped Multi-scale Deformable Convolution (GMDC)}. This approach effectively controls computational complexity through channel dimension grouping while employing multi-scale deformable kernels to adaptively capture hierarchical features, which is inspired by the hierarchy~\cite{hierarchy} of the HVS.

On the other hand, our method draws inspiration from sub-pixel analysis~\cite{espcn} known for its assumption that there is even finer microscopic information between macroscopic physical pixels, which has not been explored in the SR FR-IQA task. Building upon this foundation, we assume that there exists finer distortion information between existing artifacts; thus, we introduce the concept of \textbf{Sub-information Excitation Convolution (SubEC)}. This innovation leverages the mixed attention mechanism through the synergistic integration of sub-pixel and sub-channel information. 

We will make our code publicly available at: \url{https://github.com/Lighting-YXLI/PBAN}. To sum up, our contributions are listed as follows:

1) We propose a Perception-oriented Bidirectional Attention Network (PBAN) tailored for SR FR-IQA. At its core, PBAN introduces a Bi-Atten that aligns SR and reference features in both HR→SR and SR→HR directions. Unlike conventional attention mechanisms relying on pooling, it employs feature interaction via matrix multiplication to preserve fine-grained distortion cues introduced by SR algorithms.

2) We design a GMDC inspired by the hierarchical nature of the human visual system. It utilizes a grouped convolution strategy combined with multi-scale deformable kernels, enabling efficient modeling of hierarchical distortions while ensuring cross-channel information interaction.

3) We develop a SubEC that incorporates learnable upsampling to extract sub-pixel and sub-channel cues. By leveraging hybrid attention within spatial and channel dimensions, it effectively extracts finer perceptual signals of SR distortions.
\section{Related Work}
\subsection{Common Full Reference Image Quality Assessment}
When original reference images are available, full-reference image quality assessment (FR-IQA) methods can be developed, which estimate image quality by comparing distorted images with the corresponding reference images. The earliest FR-IQA metrics include PSNR and MSE. Later, SSIM~\cite{ssim} combined attributes such as structure, brightness, and contrast to design a metric that is more in line with the HVS. Further variants like multi-scale SSIM (MS-SSIM)~\cite{msssim}, complex wavelet SSIM (CW-SSIM)~\cite{cwssim}, quaternion structural similarity (QSSIM) ~\cite{qssim}, and gradient magnitude similarity deviation (GMSD)~\cite{gmsd} were proposed to better align with the HVS. However, these methods are not always as effective as deep learning-based methods.

Recent advances in common FR-IQA have witnessed a transition from traditional signal fidelity-based measures to deep learning-driven methods. Early deep architectures, such as WaDIQaM~\cite{bosse}, proposed unified models that jointly predict local quality and weights via convolutional networks, laying the groundwork for end-to-end FR-IQA learning. Subsequent works expanded this paradigm to multi-task frameworks, integrating distortion type classification alongside quality regression through gated feature masking strategies~\cite{MGCN}. Structural understanding was further enhanced by SPSIM~\cite{spsim}, which introduced perceptually meaningful superpixel-based similarity indices. The emergence of perceptual metrics based on deep features, as discussed in~\cite{Zhang_2018_CVPR}, revealed that pre-trained deep networks such as VGG inherently capture human visual preferences, outperforming handcrafted metrics~\cite{ssim,msssim} by large margins. To address fine-grained compression artifacts, FG-IQA~\cite{fgiqa} proposed a structure-texture difference-based method with superior discrimination at subtle quality levels. Panoramic IQA was explored in MFAN~\cite{mfan} via multi-projection attention fusion networks, adaptable to both FR and NR settings. Robustness to spatial misalignment was addressed by DOSS~\cite{doss}, leveraging order statistics of deep features to tolerate shift-induced inconsistencies. Graph-based approaches like GRIDS~\cite{grids} introduced graph representations and clique-based distribution comparisons to model perceptual distances. Most recently, DMM~\cite{dmm} tackled perceptual bias in deep feature spaces by introducing debiased mappings via SVD, balancing separability and compactness in quality estimation. However, these FR-IQA approaches are primarily designed for degradative distortions such as compression, noise, and blur, and may not generalize well to the artifacts often introduced by super-resolution algorithms.
\subsection{Super-Resolution Image Quality Assessment}
Recently, several handcrafted FR-IQA methods have been specifically designed for super-resolution (SR) images, such as the structure–texture decomposition-based SIS~\cite{sis}, the structural fidelity versus statistical naturalness (SFSN) method~\cite{sfsn}, and the deterministic and statistical fidelity-based SRIF~\cite{srif}. However, deep learning-based FR-IQA models tailored for SR images remain largely underexplored.

Current deep learning-based SR IQA research primarily focuses on no-reference scenarios. For example, DeepSRQ~\cite{deepsqr} adopts a two-stream CNN to separately extract structural and textural features, while HLSRIQA~\cite{HLSRIQA} exploits high- and low-frequency maps to better capture SR-specific artifacts. KLTSRQA~\cite{kltsrqa} evaluates the quality of SR images in a no-reference manner based on the Karhunen-Loéve Transform. EK-SR-IQA~\cite{eksriqa} introduces a semi-supervised knowledge distillation strategy to improve quality prediction. More recently, attention mechanisms have been widely applied to further improve NR-IQA performance. JCSAN~\cite{jcsan} incorporates joint channel and spatial attention to extract more perceptually relevant features, $C^2$MT~\cite{C2MT} employs a multi-task transformer to jointly predict SR image quality and classify SR algorithms, leveraging contrastive learning and pseudo-label refinement for improved perceptual representation, and TADSRNet~\cite{tadsrnet} employs a triple-attention strategy to emphasize key SR image regions across multiple dimensions.

Despite these advances, a notable gap remains in developing deep learning-based full-reference SR IQA methods. To address this, we propose the Perception-oriented Bidirectional Attention Network.
\begin{figure*}[t]
    \centering
    \includegraphics[scale=0.36]{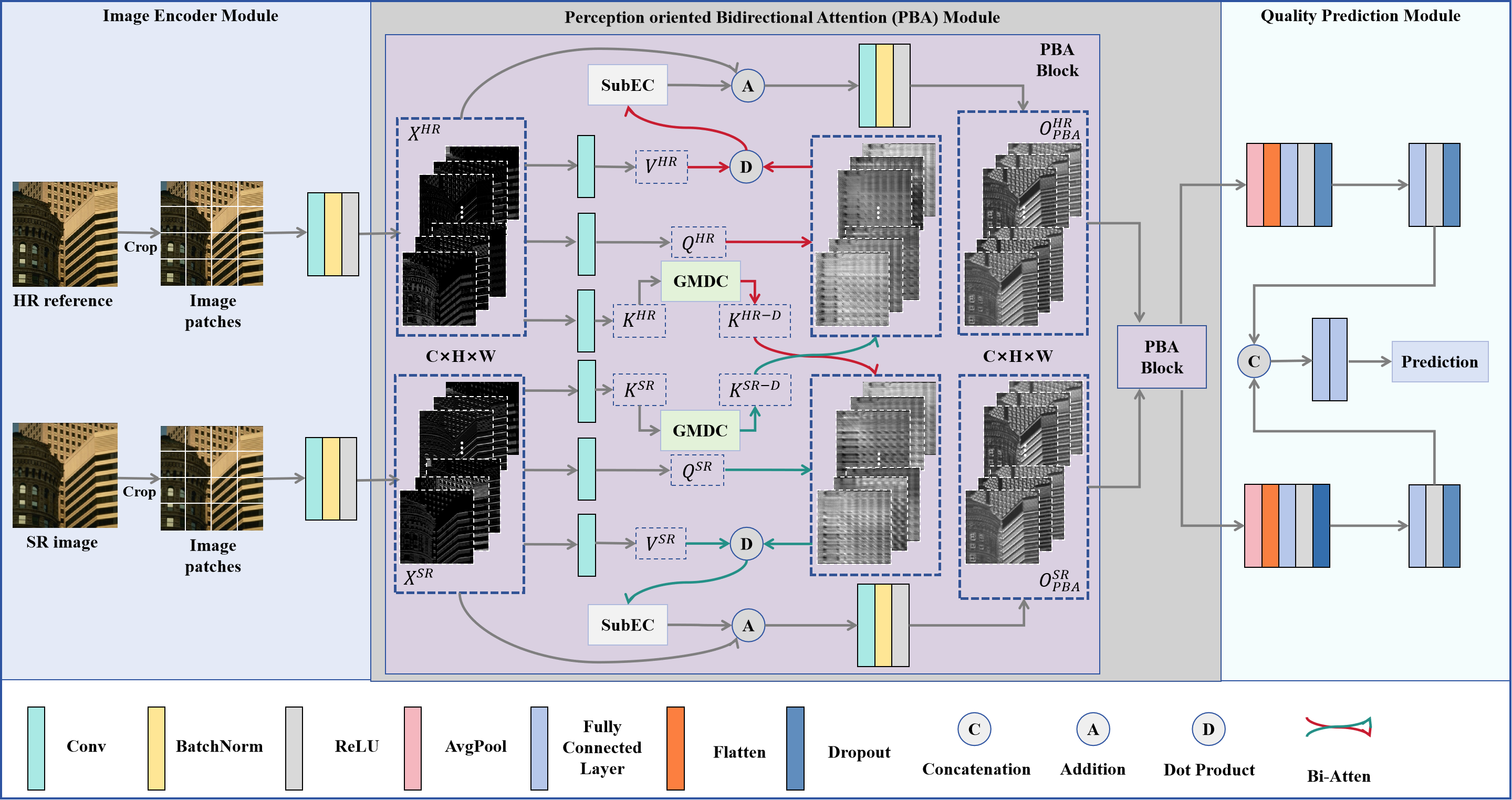}
    \caption{The overall framework of PBAN. In the PBA Module, $X^{HR}, X^{SR}$ are input feature maps, $O^{HR}_{PBA}, O^{SR}_{PBA}$ are output feature maps. $C, H, W$ are the dimensions of channel, height, and width of the feature map, respectively. \textbf{Bi-Atten} takes “K” from GMDC to compute bidirectional attention. In Bi-Atten, $Q, K, V$ represent “query”, “key”, “value” of the attention mechanism. \textbf{SubEC} models SR distortions on a finer level.}
    \label{2}
    \end{figure*}
    
\subsection{Applications of the Deformable Convolution and Sub-pixel Convolution}
Deformable convolution, initially introduced in Deformable ConvNets~\cite{dai_deformable_2017} and later improved in More Deformable ConvNets~\cite{zhu_deformable_2019}, is a technique that incorporates adaptive sizes and shapes of convolution kernels by adding learnable offsets to sampling points. It has proven to be effective in enhancing spatially dense prediction and has found applications in various low-level vision tasks like image super-resolution~\cite{deformsr} and video deblurring~\cite{edvr}. Recently, researchers have explored the use of deformable convolution in IQA to achieve performance improvements. Some notable works in this direction include RADN~\cite{radn} and DQM-IQA~\cite{dqm-iqa}. RADN~\cite{radn} utilized deformable convolutions to better leverage information from reference images, thereby increasing sensitivity to errors and misalignments in distorted images. DQM-IQA~\cite{dqm-iqa} employed deformable convolutions to extract perceptual features. However, the high computational complexity of deformable convolutions restricts the use of multi-scale convolution kernels in these methods, resulting in limitations in capturing hierarchical information.

In the field of camera imaging, sub-pixels refer to the micro-pixels that cannot be detected between macro-pixels due to limitations in the imaging capability of the camera's photosensitive element. Inspired by the concept of sub-pixels, ESPCN~\cite{espcn} introduced the concept of sub-pixel convolution for efficient image SR. Instead of padding or interpolation, it utilizes pixel shuffle to achieve image upsampling. Sub-pixel convolution aligns well with the human visual perception of SR images and has been widely applied in tasks such as image reconstruction~\cite{subpixelreconstruction} and pan-sharpening~\cite{subpixelms}.

Inspired by the aforementioned approaches, we propose Grouped Multi-scale Deformable Convolution and Sub-information Excitation Convolution to enhance perceptual SR distortion modeling.

\section{Proposed SR IQA Method}
In this section, we present the overall framework of our proposed deep learning-based SR FR-IQA method, called \textbf{Perception-oriented Bidirectional Attention Network (PBAN)}, which consists of three modules, namely the Image Encoder Module, Perception-oriented Bidirectional Attention (PBA) Module, and Quality Prediction Module. As illustrated in Fig.~\ref{2}, the network takes a HR reference and SR image pair as input. The main component of our method is the PBA Module, which is the stack of PBA blocks. Each block has three key components: Grouped Multi-scale Deformable Convolution (GMDC), Bidirectional Attention (Bi-Atten), and Sub-information Excitation Convolution (SubEC). Specifically,  we design Bi-Atten to comprehensively assess the enhanced distortions of SR images by bi-directional feature spaces alignment (HR→SR and SR→HR). Before calculating the attention map, we propose GMDC to provide the perception of distorted regions. After obtaining the Bi-Atten map, we propose the SubEC to further refine the perception of distortion to a finer level. The design details are as follows.

\subsection{Image Encoder Module}
Given a pair of input images (i.e., SR image and the corresponding original HR reference), we first crop the images into non-overlapping patches. Following the settings of DeepSRQ~\cite{deepsqr}, the patch size is set to $32\times32$ in our experiments. The image patches and their corresponding reference patches are then fed into a stack of a convolutional layer with $3\times3$ kernels, a batch normalization layer, and an activation function (i.e., ReLU) for discriminative feature extraction. 

\subsection{Perception-oriented Bidirectional Attention (PBA) Module}
The feature maps of both branches are then input into the PBA Module, which pass through \textbf{Bi-Atten} and \textbf{SubEC} in sequence.

\begin{figure*}[t]
\centering
\includegraphics[scale=0.52]{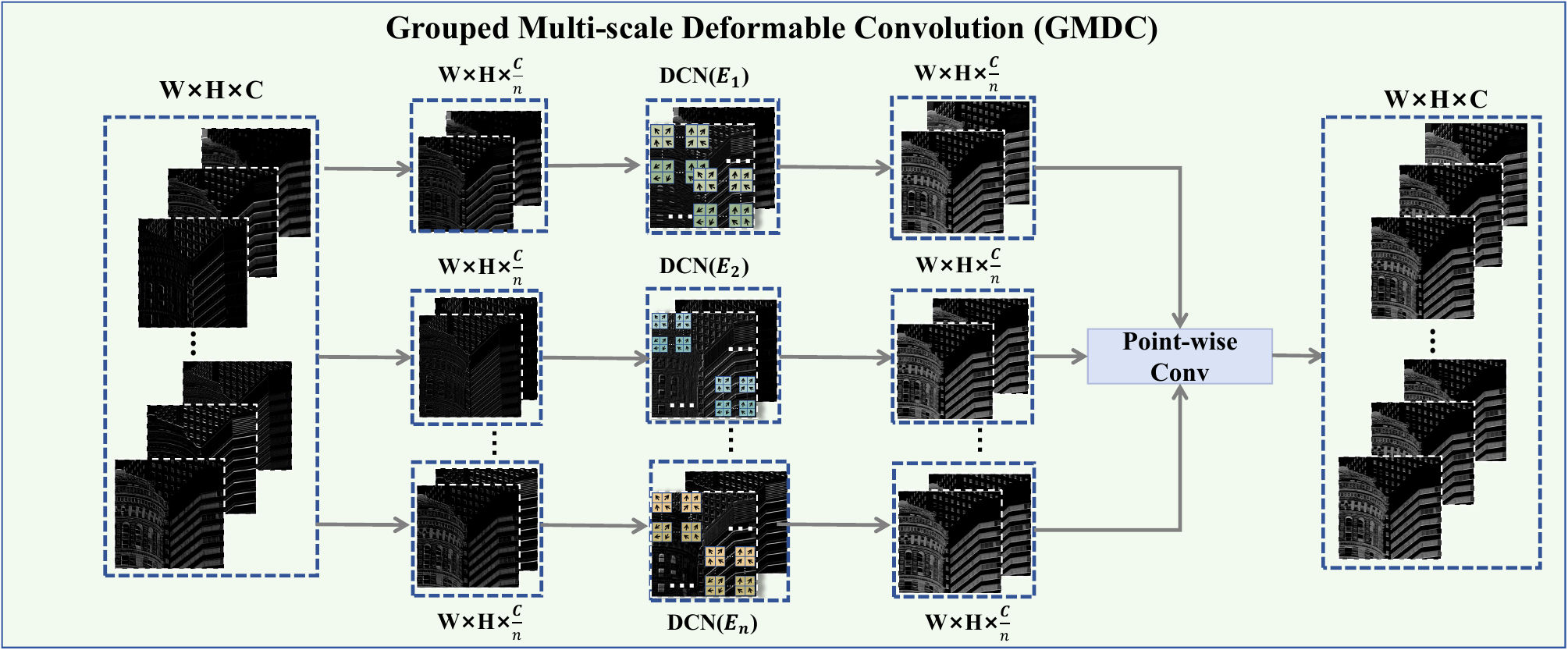}
    \caption{The framework of \textbf{GMDC}. Given an input feature map of ($W\times H\times C$), where $C$ is the number of channels, $H, W$ are the height and width. It is divided into “$n$” separate groups. Each group contains “($\frac{C}{n}$)” channels, and is then fed into deformable convolution (i.e., DCN) with $E_{i}\times E_{i}, i=1,\dots,n$ kernel sizes.  $E_{i}$ can be set to multi-scale. The final Point-wise Convolution is utilized to provide the interaction between groups.}
    \label{3}
\end{figure*}

\subsubsection{Bidirectional Attention}
Bi-Atten is an improvement upon cross-attention~\cite{crossatten} specifically for SR FR-IQA. Cross attention is commonly used for multi-modal data (such as text and images), assuming there are inputs $X_{1}, X_{2}$ from two sources. Cross attention is calculated using $Q (Query)$, $K (Key)$, and $V (Value) $ matrices to find information dependencies between cross-modal data. Meanwhile, due to the inconsistency of data between different modalities and in order to ensure contextual consistency between $K\&V$, it is necessary to ensure that $K\&V$ are homologous. The cross-attention is first obtained through linear layers:
\begin{equation}
\small
\begin{aligned}
&Q_{1},Q_{2}=W^{Q}(X_{1},X_{2}), \\
&K_{1},K_{2}=W^{K}(X_{1},X_{2}), \\
&V_{1},V_{2}=W^{V}(X_{1},X_{2}), \\
&D_{1},D_{2}=Var(Q_{1} K_{2}^{T}),Var(Q_{2} K_{1}^{T}),
\end{aligned}
\label{eq1}
\end{equation}
where $W$ refers to the fully connected linear layers. $D$ represents the variance of the dot product of $Q$ and $K$. Then the cross attention is calculated as follows:
\begin{equation}
\small
\begin{aligned}
&\operatorname{Cross-Attention}(X_{1})=\operatorname{Softmax}\left(\frac{Q_{1} K_{2}^{\top}}{\sqrt{D_{1}}}\right) V_{2}, \\
&\operatorname{Cross-Attention}(X_{2})=\operatorname{Softmax}\left(\frac{Q_{2} K_{1}^{\top}}{\sqrt{D_{2}}}\right) V_{1}. \\
\end{aligned}
\label{eq2}
\end{equation}
Here, the cross attention reflects the relevance between each element from one source $X_{1}$ and elements from the other source $X_{2}$, thereby achieving effective attention between multi-model data.

To comprehensively leverage the reference information of SR images, we provide a concise overview of the baseline, as illustrated in Fig. \ref{2}. Inspired by the “generate distortion" and “evaluate distortion" processes, we propose the dual-branch Bi-Atten. For the “generate distortion" process, the upper branch uses HR reference images as input. Then, the feature space of HR references is progressively aligned to that of SR images. This is achieved by multiplying the HR feature matrix with the SR feature matrix, where each row of the former is weighted by each column of the latter, resulting in an attention map within the SR feature space. In the “evaluate distortion" process, human evaluators typically compare SR images to corresponding references or real scenes to assess distortion levels. Considering this, the lower branch takes SR images as input and incrementally aligns them to HR reference space, effectively measuring the degree of misalignment.

The strong contextual relationship between SR images and corresponding HR references is quite different from cross-model data. \textbf{Therefore, maintaining ``$K\&V$" homology is not necessary for SR FR-IQA, but rather ``$Q\&V$" homology can emphasize the subtle differences (e.g., distortion) between SR images and corresponding HR references.} So the proposed Bi-Atten utilizes the visual information from one branch as “query \& value" and information from the other branch as “key" to achieve attention on distortion, which is different from generic cross attention.

We use $3\times3$ convolutional layers to obtain the $Q$, $K$, and $V$ matrices:
\begin{equation}
\small
\begin{aligned}
&{Q^{HR},K^{HR},V^{HR}}=Conv(X^{HR}), \\
&{Q^{SR},K^{SR},V^{SR}}=Conv(X^{SR}),
\end{aligned}
\label{eq3}
\end{equation}
where $X^{HR}, X^{SR} \in R^{C\times 32\times32}$ are the SR image patch and its corresponding HR reference patch. $C$ is the channel dimension.
\begin{figure*}[t]
\centering
\includegraphics[scale=0.265]{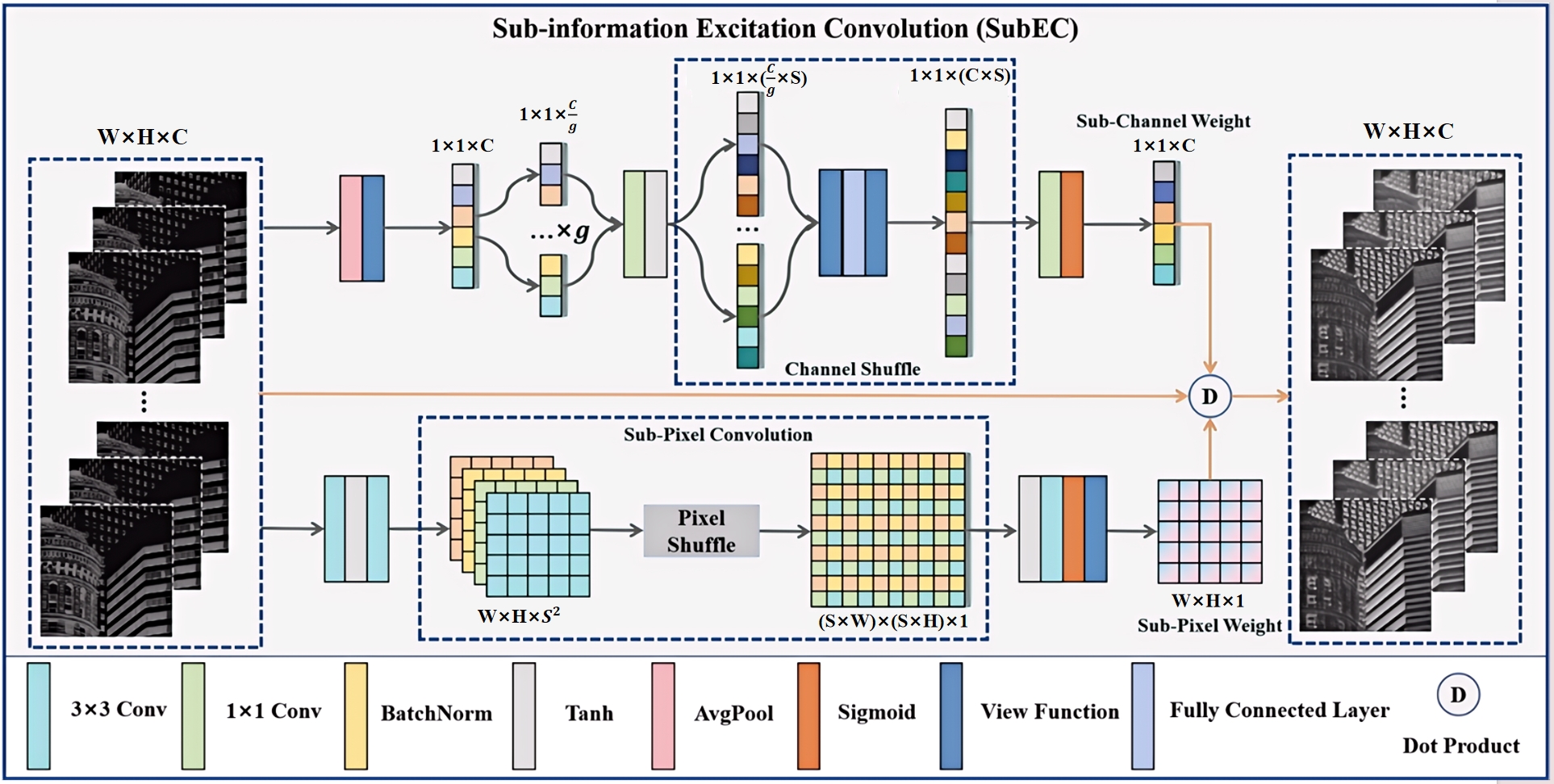}
    \caption{The framework of \textbf{SubEC}. It is a three-branch architecture, including \textbf{Sub-Channel Weight} branch (i.e., the top branch), \textbf{Sub-Pixel Weight} branch (i.e., the bottom branch), and the identity shortcut branch (i.e., the mid one). The “View Function" refers to PyTorch’s tensor.view() operation, which reshapes the tensor without altering its underlying data. For example, the first View Function in the Sub-Channel Weight branch reshapes the feature map from $[B, C]$ to $[B, C, 1, 1]$, where B and C denote the batch size and number of channels, respectively. The output feature map is obtained by multiplying the input feature map by these two weights. $S$ is the magnification of the channel and pixel dimensions, which is set to 2 in our experiment. }
    \label{4}
\end{figure*}

Then the two branches exchange $K^{HR}$ and $K^{SR}$ for information interaction. Within Bi-Atten, we further propose \textbf{GMDC}, which enhances the adaptability of spatial dense predictions and consequently strengthens the perception of visual attention. The detailed formulation is introduced as follows:

In standard convolution, the convolution operation can be calculated as follows:
\begin{equation}
\small
F\left(p_{0}\right)=\sum_{p_{n} \in r} w\left(p_{n}\right) \cdot y\left(p_{0}+p_{n}\right),
\label{eq4}
\end{equation}
where $p_{0}$ is a pixel of the output feature map $y$, $r$ denotes the regular grid of its corresponding kernel, $p_{n}$ enumerates the relative locations of sampling points in $r$. $w$ denotes the weight for $p_{n}$. Deformable convolution (DCN) utilized in IQA methods~\cite{radn,dqm-iqa} adds vertical and horizontal offsets $\Delta p_{n}$ to each sampling point of a convolution kernel:
\begin{equation}
\small
F\left(p_{0}\right)=\sum_{p_{n} \in r} w\left(p_{n}\right) \cdot y\left(p_{0}+p_{n}+\Delta p_{n}\right).
\label{eq5}
\end{equation}
The offsets are learned from the preceding feature map $y$ via a convolution layer.  
However, the addition of offsets in DCN introduces a significant number of extra parameters, making it impractical to use larger convolution kernels and limiting its learning capacity. Instead, group convolution~\cite{resnext} can obtain more feature maps while maintaining the same parameter quantity as standard convolution, which can provide the feature extraction capability. Thus, the proposed GMDC enables hierarchical modeling by employing multi-scale kernel sizes in parallel while effectively controlling the number of parameters through the grouped structure. As shown in Fig. \ref{3}, given inputs with the size of ($W\times H\times C$), where $C$ is the number of channels, $H, W$ are the height and width of the input. We divide channels into $n$ groups (i.e., each group has ($\frac{C}{n}$) channels). Different groups undergo DCN operations independently. To capture information at multiple scales,  we apply multi-sized convolution kernels with edge length $E_{i}, i=1,\dots,n$ to different groups. Take one convolutional layer as an example, the parameters for standard DCN are $3\times C'\times C\times E^{2}$, and the parameters for the proposed GMDC are $3\times C'\times C\times\frac{E_{1}^{2}+\dots+E_{n}^{2}}{n}$, where the $C'$ refers to the channel of the output features, $E, E_i, i=1,\dots, N$ refer to kernel sizes. When $\tfrac{1}{n}\sum E_i^{2} \approx E^{2}$, GMDC has the same order of parameters as DCN, but by employing various kernel sizes, it introduces hierarchical modeling. The ablation study is detailed in Table \ref{t3}. After going through GMDC, $K^{HR}$ and $K^{SR}$ are swapped and involved in the computation of the Bi-Atten map in the other branch:
\begin{equation}
\small
\begin{aligned}
&K^{HR-D},K^{SR-D}=GMDC(K^{HR},K^{SR}),\\
&D^{HR}=Var(Q^{HR} {(K^{SR-D})}^{T}), \\
&D^{SR}=Var(Q^{SR} {(K^{HR-D})}^{T}),
\end{aligned}
\label{eq6}
\end{equation}
where $D^{HR}$ and $D^{SR}$ are the variance of the dot product of $Q^{HR},K^{SR-D}$ and $Q^{SR}, K^{HR-D}$, respectively. Note that the proposed Bi-Atten is different from cross attention and common spatial/channel attention, which are further analyzed in Section \ref{cross} and Section \ref{spatialchannel}, respectively. Then the Bidirectional Attention maps are calculated as:
\begin{align}
\small
&\operatorname{Bi-Atten^{HR}}(Q^{HR}, K^{SR-D}, V^{HR})\notag \\ 
&=\operatorname{Softmax}\left(\frac{Q^{HR} {K^{SR-D}}^{\top}}{\sqrt{D^{HR}}}\right) V^{HR},
\end{align}
\begin{align}
\small
&\operatorname{Bi-Atten^{SR}}(Q^{SR}, K^{HR-D}, V^{SR})\\\notag
&=\operatorname{Softmax}\left(\frac{Q^{SR} {K^{HR-D}}^{\top}}{\sqrt{D^{SR}}}\right) V^{SR}.
\end{align}
\subsubsection{Sub-information Excitation Convolution}
Consequently, we introduce SubEC to extract finer distortion information in both the spatial (pixel) and the channel dimensions of the obtained Bi-Atten feature maps. 

Based on the concept of sub-pixel~\cite{espcn}, we assume that micro-distorted information exists in the sub-pixels between pixels in the feature maps. Simultaneously, it also exists in sub-channels, which are assumed to be mined between the existing channels. Therefore, we mine sub-information through super-resolution of both the channel and spatial dimensions.
\begin{figure*}[t]
\centering
\includegraphics[scale=0.175]{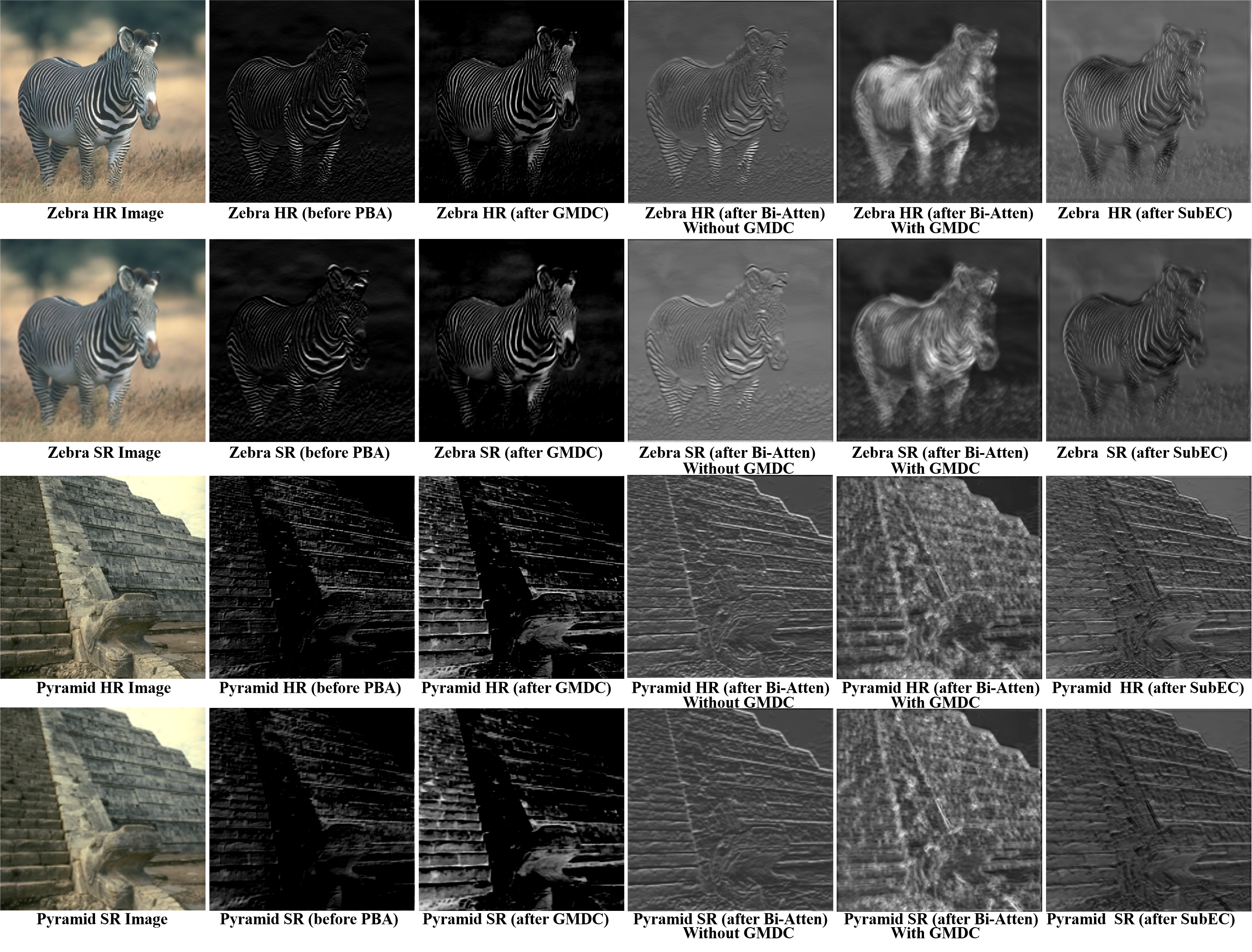}
    \caption{Visualization comparisons of feature maps regarding the proposed PBA block. \textbf{Zebra/Pyramid HR image} and \textbf{Zebra/Pyramid SR image} are HR reference and SR image, respectively. The remaining images are feature maps before and after the first PBA block and its components in two branches.}
    \label{5}
\end{figure*}
\begin{figure}[t]
\centering   
\includegraphics[scale=0.51]{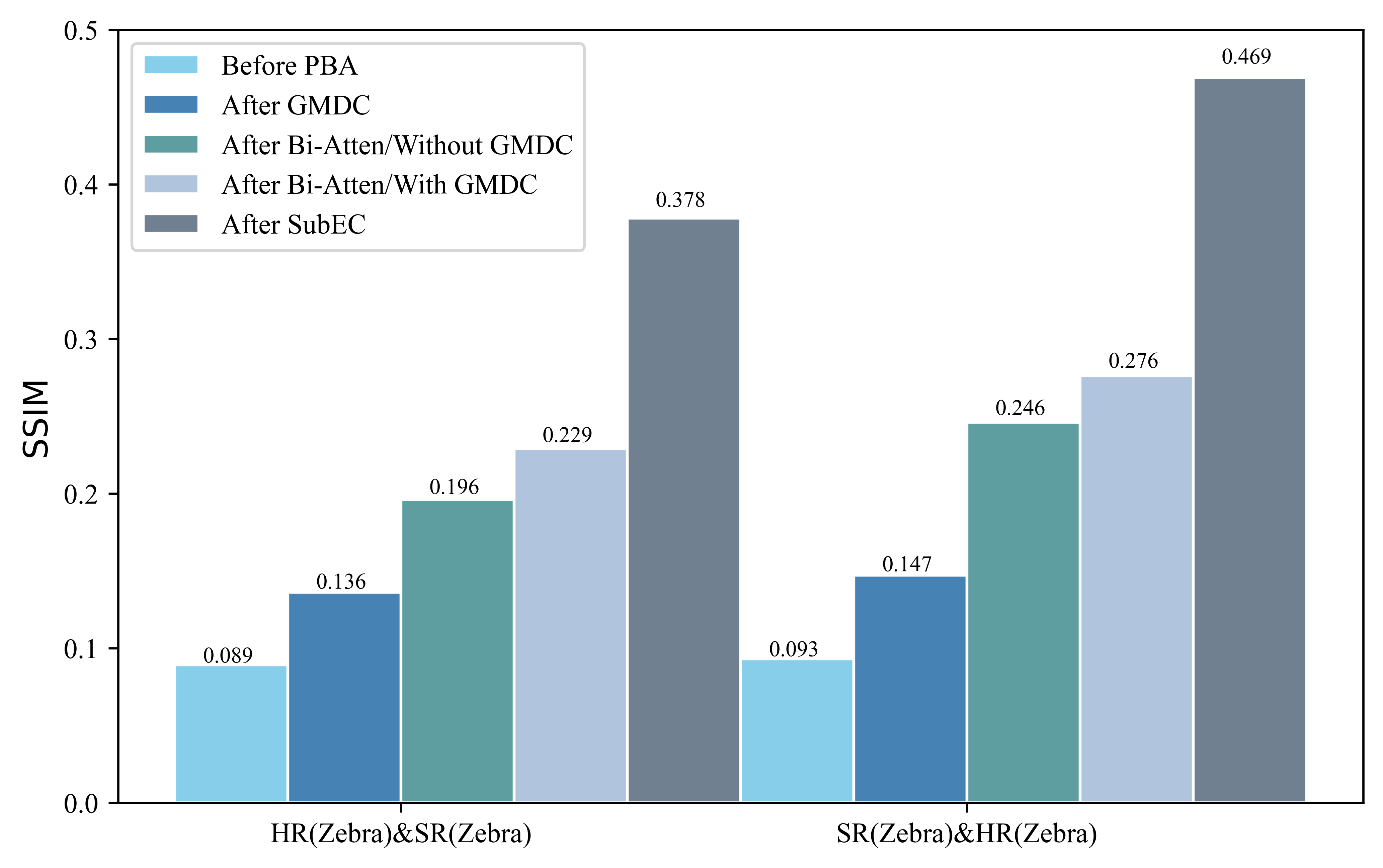}
    \caption{The structure similarity (i.e., SSIM) of “$X_{1}\& X_{2}$”, where $X_{1}$ is the initial HR reference or SR image, $X_{2}$ is the \textbf{feature map} before or after one PBA block and its components (i.e., GMDC located in Bi-Atten, Bi-Atten w/o GMDC, Bi-Atten with GMDC, and SubEC). Note that the higher the SSIM, the more similar the two images are.}
    \label{6}
\end{figure}

The proposed SubEC framework, illustrated in Fig. \ref{4}, takes an input feature map of size ($W\times H\times C$) and generates the Sub-Channel Weight and Sub-Pixel Weight through separate branches.

In the Sub-Pixel Weight branch, we employ sub-pixel convolution~\cite{espcn} for pixel-level super-resolution. With an upscaling factor of $S$, each pixel in the feature map of size ($W\times H$) is assigned weights. Consequently, the final size of Sub-Pixel Weight should be ($W\times H\times 1$). Sub-pixel convolution upsamples a single pixel into a square matrix of size ($S\times S$). To achieve this, we first reduce the channel dimension to $S^{2}$ and then use sub-pixel convolution to shuffle the pixels, as illustrated in Fig. \ref{4}. This process generates an upsampled feature map of size ($(S\times W)\times (S\times H)\times1$). To avoid losing channel information when reducing the number of channels from $C$ to $S^{2}$, we gradually decrease the dimension using two $3\times3$ convolutional layers. Finally, the upsampled feature map is downsampled to ($W\times H\times1$) to obtain the Sub-Pixel Weight.

In the Sub-Channel Weight branch, we amplify the channel dimension to capture micro-information hidden within sub-channels. Since the final weights are applied to each input channel, the size of Sub-Channel Weight needs to be ($1\times 1\times C$). Given an input feature map, we first reduce the spatial dimension from ($W\times H$) to ($1\times 1$) using adaptive average pooling. Then, we perform super-resolution on the channel dimension by a factor of $S$. To provide computational efficiency, we employ group convolution~\cite{resnext} to divide the feature map into $g$ groups, with each group having a size of ($1\times1\times\frac{C}{g}$). Subsequently, a $1\times1$ convolution is applied to expand the channels by S times. To restore the interdependencies among channels lost during grouping, we utilize the channel shuffle operation~\cite{shufflenet}, resulting in a feature map of size ($1\times1\times(C\times S)$). Finally, we compress the channels using a $1\times1$ convolution to obtain the Sub-Channel Weight.
In summary, the output feature map $O_{SubEC}$ can be expressed as:
\begin{equation}
\small
O_{SubEC}=I_{SubEC}\times W_{Sub-Channel}\times W_{Sub-Pixel},
\label{eq8}
\end{equation}
where $I_{SubEC}$ is the input of SubEC. Taking into account that the Identity Shortcut~\cite{resnet} has been proven to effectively alleviate model overfitting issues, we use the Shortcut as the main architecture:
\begin{equation}
\small
\begin{aligned}
&O^{HR}_{PBA}=O_{SubEC}^{HR}+X^{HR},\\
&O^{SR}_{PBA}=O_{SubEC}^{SR}+X^{SR},
\end{aligned}
\label{eq9}
\end{equation}
where $O^{HR}_{PBA},O^{SR}_{PBA}$ are output feature maps of PBA block.
\subsection{Quality Prediction Module}
After passing through the PBA Module, feature maps are fed into the \textbf{Quality Prediction Module}, where the feature maps of both branches are flattened and concatenated:
\begin{equation}
\small
\begin{aligned}
&O^{HR}=ReLU(L(F(AvgPool(O^{HR}_{PBA})))),\\
&O^{SR}=ReLU(L(F(AvgPool(O^{SR}_{PBA})))),\\
&O^{HR}=Dropout(ReLU(L(Dropout(O^{HR})))),\\
&O^{SR}=Dropout(ReLU(L(Dropout(O^{SR})))),
\end{aligned}
\label{eq10}
\end{equation}
where the Average Pooling layer is utilized to downsample $O^{HR}_{PBA}$ and $O^{SR}_{PBA}$ to prevent the input dimension of the fully connected layer from being too large. $F$ is the flattened layer and $L$ is the fully connected layer. These Linear layers gradually reduce the feature dimension from 1,024 to 1, resulting in the final prediction score.

Afterward, the perceptual quality predictions are ultimately obtained:
\begin{equation}
\small
\begin{aligned}
&O=Concate(O^{HR},O^{SR}),\\
&Prediction=Linear(Linear(O)).
\end{aligned}
\label{eq11}
\end{equation}

Finally, following the protocol of DeepSRQ~\cite{deepsqr}, we adopt the MSE loss to provide guidance in the network optimization process:
\begin{equation}
\small
Loss = \frac{1}{B}\sum_{i=1}^{B}(Prediction_{i}-MOS_{i})^{2},
\end{equation}
where $B$ is the batch size of the input data.
\section{Visualization}
To demonstrate the effectiveness of the proposed PBAN and its components in enhancing visual attention to distortions in SR images, Fig. \ref{5} presents the intermediate feature maps of the two branches. The branches take the HR reference image and the SR image of “zebra” and “pyramid” as inputs, respectively. As shown, the initial feature maps extracted by the image encoder module (“image before PBA”) capture limited distortion information. After processing with GMDC, the feature maps of $K^{HR-D}$ and $K^{SR-D}$ exhibit stronger responses to the “zebra” and “pyramid” regions, indicating that multi-scale adaptive sampling effectively facilitates spatially dense prediction. Subsequently, Bi-Atten is applied to generate bi-directional attention maps between the two branches, which significantly enhance the focus on artifact regions. Finally, after passing through SubEC and producing the output of one PBA block, the distortion details are further refined.

To visually demonstrate the perceptual gain of GMDC, we compared the feature maps of Bi-Atten with and without GMDC, and observed that the version with GMDC focuses more on distortions (e.g., zebra textures, brick structures).

Furthermore, the PBA module is designed to reflect the processes of SR image distortion “generation” and “evaluation,” thereby enabling more comprehensive predictions consistent with human perception. During iterative learning, the feature maps of the branch with the HR reference as input are expected to gradually approximate those of the SR image, while the branch with the SR image as input should, conversely, approach the HR reference. To quantify this approximation, we compute similarity metrics (i.e., SSIM) on the feature maps of “zebra,” as illustrated in Fig. \ref{6}.

The results show that the feature maps of the SR image achieve a notable SSIM improvement with respect to the HR reference after receiving visual attention from the first PBA block. Similarly, the feature maps of the HR reference also show a marked SSIM improvement with respect to the SR image. These findings suggest that our method dynamically enhances perceptual sensitivity to distortions by allowing the HR and SR representations to progressively align with each other.
\section{Experiments}
\begin{table*}[t]
\small
  \centering
  \caption{Performance comparisons on the QADS~\cite{sis}, CVIU~\cite{cviu}, and Waterloo~\cite{waterloo} quality databases, where the best test performance values of FR and NR are in \textcolor{red}{red} and \textcolor{blue}{blue}, respectively. All DNN-based IQA methods are independently trained on each individual database.}
  \resizebox{\linewidth}{!}{
    \begin{tabular}{c|c|cccc|cccc|cccc}
    \toprule
          &       & \multicolumn{4}{c|}{QADS}     & \multicolumn{4}{c|}{CVIU}  & \multicolumn{4}{c}{Waterloo}\\
    \midrule
    Types & Model & SRCC $\uparrow$ & KRCC $\uparrow$  & PLCC $\uparrow$ & RMSE $\downarrow$  & SRCC $\uparrow$  & KRCC $\uparrow$ & PLCC $\uparrow$  & RMSE $\downarrow$ & SRCC $\uparrow$  & KRCC $\uparrow$  & PLCC $\uparrow$  & RMSE $\downarrow$\\
    \midrule
    \multirow{8}[2]{*}{FR-IQA} & PSNR  & 0.354 & 0.244 & 0.390 & 0.253 & 0.566 & 0.394 & 0.578 & 1.962 &0.632 & 0.442  &0.630  &2.002 \\
          & SSIM~\cite{ssim}  & 0.529 & 0.369 & 0.533 & 0.233 & 0.629 & 0.443 & 0.650 & 1.828 &0.613  &0.431  &0.621  & 2.022\\
          & MS-SSIM~\cite{msssim} & 0.717 & 0.530 & 0.724 & 0.190 & 0.805 & 0.601 & 0.811 & 1.405 &0.825 &0.623 &0.837 &1.411 \\
          & CW-SSIM~\cite{cwssim} & 0.326 & 0.228 & 0.379 & 0.254 & 0.759 & 0.541 & 0.754 & 1.579 & 0.863 & 0.666  &0.906 &1.094\\
          & GMSD~\cite{gmsd}  & 0.765 & 0.569 & 0.775 & 0.174 & 0.847 & 0.650 & 0.850 & 1.267 &0.797 &0.592 &0.811 &1.509 \\
          & WaDIQaM~\cite{bosse} & 0.871 & \makebox[1em][c]{\textemdash} & 0.887 & 0.128 & 0.872 & \makebox[1em][c]{\textemdash} & 0.886 & 1.304 & \makebox[1em][c]{\textemdash} &\makebox[1em][c]{\textemdash} &\makebox[1em][c]{\textemdash} &\makebox[1em][c]{\textemdash}\\
          & LPIPS~\cite{LPIPS}& 0.881 & \makebox[1em][c]{\textemdash} & 0.873 & 0.129 & 0.849 & \makebox[1em][c]{\textemdash} & 0.852 & 1.313 & \makebox[1em][c]{\textemdash}& \makebox[1em][c]{\textemdash}&\makebox[1em][c]{\textemdash} &\makebox[1em][c]{\textemdash}\\
    \midrule
    \multirow{5}[2]{*}{NR-IQA} & NIQE~\cite{niqe}  & 0.398 & 0.279 & 0.404 & 0.251 & 0.653 & 0.478 & 0.666 & 1.794 &0.626 &0.465 &0.672 &1.911 \\
          & LPSI~\cite{lpsi}  & 0.408 & 0.289 & 0.422 & 0.249 & 0.488 & 0.350 & 0.537 & 2.027& 0.667 & 0.464 &0.701 &1.840 \\
          & MetaIQA~\cite{metaiqa}& 0.826 & \makebox[1em][c]{\textemdash} & 0.790 & 0.178 & 0.720 & \makebox[1em][c]{\textemdash} & 0.746 & 1.718&\makebox[1em][c]{\textemdash} &\makebox[1em][c]{\textemdash} &\makebox[1em][c]{\textemdash} &\makebox[1em][c]{\textemdash} \\
          & HyperIQA~\cite{hyperiqa} & 0.954 & 0.815 & 0.957 & 0.099 & 0.933 & 0.772 & 0.928 & 1.017&\makebox[1em][c]{\textemdash} &\makebox[1em][c]{\textemdash} & \makebox[1em][c]{\textemdash}& \makebox[1em][c]{\textemdash}\\
          & CLIP-IQA~\cite{clipiqa} & 0.931 & 0.780 & 0.833 & \makebox[1em][c]{\textemdash} & 0.930 & 0.770 & 0.931 & 0.106 & 0.912 & 0.737 & 0.900 &\makebox[1em][c]{\textemdash} \\
          & TOPIQ~\cite{chen2024topiq} & 0.969 & 0.852 & 0.970 & 0.072 & \textcolor{blue}{0.970} & \textcolor{blue}{0.858} & \textcolor{blue}{0.971} & \textcolor{blue}{0.486} & \textcolor{blue}{0.945} &\textcolor{blue}{ 0.813} & \textcolor{blue}{0.967} &\textcolor{blue}{0.683} \\
    \midrule
    \multirow{6}[2]{*}{SR NR-IQA} & DeepSRQ~\cite{deepsqr} & 0.953 & \makebox[1em][c]{\textemdash}    & 0.956 & 0.077 & 0.921 & \makebox[1em][c]{\textemdash}    & 0.927 & 0.904&0.907 & \makebox[1em][c]{\textemdash}& 0.904 & \makebox[1em][c]{\textemdash}\\
          & HLSRIQA~\cite{HLSRIQA} & 0.961 & 0.829 & 0.950 & 0.741 & 0.948 & 0.810 & 0.948 & 0.775&\makebox[1em][c]{\textemdash} &\makebox[1em][c]{\textemdash} & \makebox[1em][c]{\textemdash}& \makebox[1em][c]{\textemdash}\\
          & EK-SR-IQA~\cite{eksriqa} & 0.963 & \makebox[1em][c]{\textemdash} & 0.966 & \makebox[1em][c]{\textemdash} & 0.953 & \makebox[1em][c]{\textemdash} & 0.951 & \makebox[1em][c]{\textemdash} &0.915 &\makebox[1em][c]{\textemdash} &  0.905 &\makebox[1em][c]{\textemdash}\\
          & JCSAN~\cite{jcsan} & 0.971 & 0.858 & 0.973 & \textcolor{blue}{0.065} & 0.949 & 0.808 & 0.957 & 0.777& \makebox[1em][c]{\textemdash}&\makebox[1em][c]{\textemdash} &\makebox[1em][c]{\textemdash} &\makebox[1em][c]{\textemdash}\\
          & TADSRNet~\cite{tadsrnet} & \textcolor{blue}{0.972} & \textcolor{blue}{0.862} & \textcolor{blue}{0.974} & 0.067 & 0.952 & 0.812 & 0.959 & 0.797&\makebox[1em][c]{\textemdash} & \makebox[1em][c]{\textemdash}&\makebox[1em][c]{\textemdash} &\makebox[1em][c]{\textemdash}\\
          & $\rm C^{2}$MT~\cite{C2MT} & 0.969 & 0.854 & 0.972 & 0.092 & 0.947 & 0.796 & 0.944 & 1.180 &\makebox[1em][c]{\textemdash} & \makebox[1em][c]{\textemdash}&\makebox[1em][c]{\textemdash} &\makebox[1em][c]{\textemdash}\\
          \midrule
    \multirow{4}[2]{*}{SR FR-IQA} & SIS~\cite{sis} & 0.913 & 0.740 & 0.914 & 0.112 & 0.869 & 0.686 & 0.897 & 1.061 &0.878 &  0.677 & 0.891 &1.169\\
          & SFSN~\cite{sfsn} & 0.841 & 0.655 & 0.845 & 0.147 & 0.871 & 0.680 & 0.885 & 1.120&0.887  &0.692  &0.906 &1.093 \\
          & SRIF~\cite{srif} & 0.916 & 0.746 & 0.917 & 0.109 & 0.886 & 0.704 & 0.902 & 1.039 & 0.916 &0.730  &0.953  &0.786\\
          & Proposed PBAN & \textcolor{red}{0.986} & \textcolor{red}{0.923} & \textcolor{red}{0.987} & \textcolor{red}{0.044} & \textcolor{red}{0.978} & \textcolor{red}{0.872} & \textcolor{red}{0.981} & \textcolor{red}{0.397} &\textcolor{red}{0.965} & \textcolor{red}{0.848}&\textcolor{red}{0.979} &\textcolor{red}{0.587}\\
    \bottomrule
    \end{tabular}}
  \label{t1}%
\end{table*}%
In this section, we compare our method with many existing quality metrics on three publicly available subject-rated quality databases for SR images. Besides, ablation studies are also conducted to verify the performance of each proposed component.
\begin{figure*}[t]
\centering
\includegraphics[width=\linewidth]{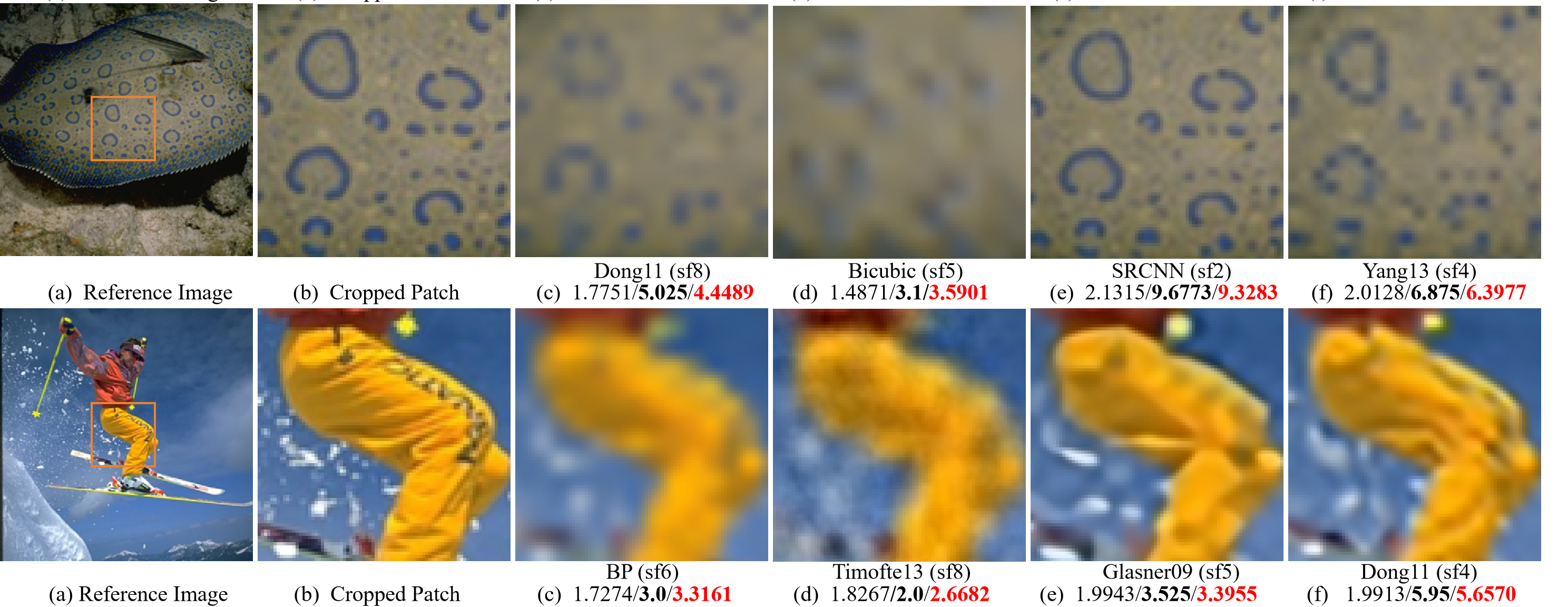}
    \caption{Examples for predicted scores on the CVIU database (test): SFSN/\textbf{PBAN}/\textcolor{red}{MOS}. The first column has HR reference images, and columns (c)-(d) are the selected SR images. All images are cropped for better visibility. In order to observe significant differences in the quality of SR images participating in the comparison, we screen images with various SR algorithms and scaling factors (i.e., sf), which are presented in detail.}
    \label{7}
\end{figure*}
\subsection{Experimental setup and implementation}
In training and testing, experiments are running on a NVIDIA DGX Station with a Tesla V100-DGXS-32GB GPU and Ubuntu 18.04 LTS. All codes are implemented in PyTorch.
\subsubsection{Databases}
We conduct experiments on QADS~\cite{sis}, CVIU~\cite{cviu}, and Waterloo~\cite{waterloo} databases. The QADS database contains 20 original HR references and 980 SR images created by 21 SR algorithms, including 4 interpolation-based, 11 dictionary-based, and 6 DNN-based SR models, with upsampling factors equaling 2, 3, and 4. Each SR image is associated with the MOS of 100 subjects. In the CVIU database, 1,620 SR images are produced by 9 SR approaches from 30 HR references. Six pairs of scaling factors and kernel widths are adopted, where a larger subsampling factor corresponds to a larger blur kernel width. Each image is rated by 50 subjects, and the mean of the median 40 scores is calculated for each image as the MOS. The Waterloo database involves 8 interpolation algorithms with 3 interpolation factors of 2, 4, and 8, respectively. 312 SR images are generated from 13 source images.

\subsubsection{Implementation details}
All databases are randomly divided into non-overlapping 80$\%$ and 20$\%$ sets, with $80\%$ of the data used for training and the remaining $20\%$ for testing. All DNN-based IQA methods are independently trained on each individual database. During training, we employ data augmentation to mitigate overfitting by dividing each SR image and its corresponding HR reference image into non-overlapping patches. Following the protocol in DeepSRQ~\cite{deepsqr}, the patch size is set to $32\times32$. Accordingly, the number of training SR–HR patch pairs is 129,360 for QADS (784 training image pairs), 194,400 for CVIU (1,296 training image pairs), and 55,800 for Waterloo (248 training image pairs). Each patch is assigned the MOS of its corresponding SR image, and the final prediction for an SR image is obtained by averaging the predictions of its patches. We utilize 5-fold cross-validation on the training set, with each fold training for 100 epochs, and ultimately test it on the test set. We use MSE loss to measure the difference between predicted scores and MOSs. The optimizer used is stochastic gradient descent (SGD), with an initial learning rate of $0.01$, momentum of $0.9$, and weight decay set to $10^{-6}$.

We adopt four commonly used evaluation criteria to compare performance, including Spearman rank-order correlation coefficient (SRCC), Kendall rank-order correlation coefficient (KRCC), Pearson linear correlation coefficient (PLCC), and root mean square error (RMSE). SRCC and PLCC/RMSE are employed to assess the monotonicity and accuracy of predictions, respectively. KRCC is used to measure the ordinal association between two measured quantities. An ideal quality metric would have SRCC, KRCC, and PLCC values close to one, and RMSE close to zero. It should be noted that a five-parameter nonlinear fitting process~\cite{srif} is applied to map the predicted qualities into a standardized scale of subjective quality labels before calculating PLCC and RMSE for different quality metrics.
\begin{table*}[htbp]
\scriptsize
  \centering
  \caption{Ablation of individual proposed components. Comparison of the test results of PBAN and variants without each component on both the QADS and CVIU databases. The best results are highlighted in bold.} 
    \begin{tabular}{ccc|cccc|cccc}
    \toprule
    \multicolumn{3}{c|}{Component} &     \multicolumn{4}{c|}{QADS}     & \multicolumn{4}{c}{CVIU} \\
    \midrule
    \multicolumn{1}{c|}{Bi-Atten} & \multicolumn{1}{c|}{GMDC} & SubEC & SRCC $\uparrow$  & KRCC $\uparrow$  & PLCC $\uparrow$  & RMSE $\downarrow$ & SRCC $\uparrow$  & KRCC $\uparrow$ & PLCC $\uparrow$ & RMSE $\downarrow$ \\
    \midrule
    \ding{53}  & \ding{53} & \ding{53} & 0.936 & 0.795 & 0.937 & 0.103 & 0.942 & 0.808 & 0.954 & 0.815 \\
    \checkmark     & \ding{53}    & \ding{53}    & 0.981 & 0.895 & 0.982 & 0.055 & 0.972 & 0.862 & 0.976 & 0.515 \\
    \checkmark & \checkmark & \ding{53} & 0.984 & 0.913 & 0.985 & 0.048 & 0.974 & 0.866 & 0.977 & 0.605\\
    \checkmark     & \ding{53}  & \checkmark    & 0.984 & 0.908 & 0.986 & 0.050 & 0.976 & 0.871 & 0.977 & 0.436 \\
    \ding{53}     & \checkmark    & \ding{53} & 0.955 & 0.827 & 0.958 & 0.082 & 0.952 &  0.816 &  0.958 &  0.773 \\
    \ding{53}     & \ding{53}    & \checkmark  & 0.961 & 0.849 & 0.962 & 0.074 &  0.966   &   0.845  &  0.969 & 0.641 \\
    \checkmark  & \checkmark & \checkmark & \textbf{0.986} & \textbf{0.923} & \textbf{0.987} & \textbf{0.044} & \textbf{0.978} & \textbf{0.872} & \textbf{0.981} & \textbf{0.397} \\
    \bottomrule
    \end{tabular}%
  \label{t2}%
\end{table*}%
\begin{table}[t]
\centering
\caption{The comparison of computational efficiency, with the fastest and second-fastest models highlighted in \textcolor{red}{red} and \textcolor{blue}{blue}, respectively. And the best and second-best performances are highlighted in \textbf{bold} and \underline{underlined}, respectively. FLOPs are calculated using an input tensor of shape $3 \times 512 \times 512$, and inference speed is averaged over 100 runs. All evaluations are conducted on the QADS database.}
\resizebox{\linewidth}{!}{
\begin{tabular}{c|ccc|cc}
\toprule
Model & Params/M & Flops/G & Speed/ms & SRCC $\uparrow$& PLCC $\uparrow$ \\
\midrule
WaDIQaM~\cite{bosse} & 6.287 & 30.479 & \textcolor{blue}{8.170} & 0.871 & 0.887 \\
HyperIQA~\cite{hyperiqa} & 27.375 & 107.831 & 29.980 & 0.954 & 0.957 \\
CLIP-IQA~\cite{clipiqa} &\makebox[1em][c]{\textemdash} & 61.065 & 21.100 & 0.931 & 0.833 \\
TOPIQ~\cite{chen2024topiq} & 36.039 & 50.138 & 22.140 & 0.969 & 0.970 \\
DeepSRQ~\cite{deepsqr} & \textcolor{red}{1.317} & \textcolor{red}{1.589} & \textcolor{red}{2.291} & 0.953 & 0.956 \\
TADSRNet~\cite{tadsrnet} & 2.119 & 303.076 & 87.568 & 0.972 & 0.974 \\
PFIQA~\cite{pfiqa} & 38.772 & 127.109 & 18.14 & \underline{0.982} & \underline{0.983} \\
PBAN & \textcolor{blue}{2.222} & \textcolor{blue}{26.278} & 16.995 & \textbf{0.986} & \textbf{0.987}\\
\bottomrule
\end{tabular}}
\label{complexity}
\end{table}
\subsection{Performance Comparisons}
To validate the proposed method, we compare it with state-of-the-art FR-IQA, NR-IQA, and SR IQA methods. FR-IQA methods include PSNR, SSIM~\cite{ssim}, MS-SSIM~\cite{msssim}, CW-SSIM~\cite{cwssim}, GMSD~\cite{gmsd}, WaDIQaM~\cite{bosse}, and LPIPS~\cite{LPIPS}. NR-IQA methods consist of NIQE~\cite{niqe}, LPSI~\cite{lpsi}, MetaIQA~\cite{metaiqa}, HyperIQA~\cite{hyperiqa}, CLIP-IQA~\cite{clipiqa}, and TOPIQ~\cite{chen2024topiq}. Among them, WaDIQaM, LPIPS, MetaIQA, HyperIQA, CLIP-IQA, and TOPIQ are deep learning-based models. SR IQA methods contain SIS~\cite{sis}, SFSN~\cite{sfsn}, SRIF~\cite{srif}, DeepSRQ~\cite{deepsqr}, HLSRIQA~\cite{HLSRIQA}, EK-SR-IQA~\cite{eksriqa}, JCSAN~\cite{jcsan}, $C^2$MT~\cite{C2MT}, and TADSRNet~\cite{tadsrnet}. Besides, DeepSRQ, HLSRIQA, EK-SR-IQA, JCSAN, $C^2$MT, and TADSRNet are deep learning-based methods.

The performance comparison is summarized in TABLE~\ref{t1}. It can be observed that deep learning-based IQA methods consistently outperform traditional hand-crafted methods across all three databases, validating the advantage of data-driven feature learning. In particular, advanced general-purpose NR-IQA models such as CLIP-IQA and TOPIQ achieve results comparable to SR-specific NR-IQA methods, indicating that recent large-scale pretraining and prompt-based modeling can enhance generalization. However, traditional FR-IQA models (e.g., PSNR, WaDIQaM, LPIPS) still show a clear performance gap, highlighting the necessity of designing SR-aware FR-IQA metrics that can capture enhanced distortions such as over-sharpening and hallucinated textures.

Among SR NR-IQA methods, attention-based networks like JCSAN and TADSRNet stand out, confirming the effectiveness of integrating visual saliency and semantic dependencies. However, their attention modules rely heavily on frequent pooling operations for feature aggregation, which inevitably causes the loss of fine-grained information. This characteristic may hinder their ability to capture subtle and localized enhanced distortions, such as edge over-sharpening or texture hallucination, which are critical in SR scenarios. In contrast, our PBAN adopts a bidirectional attention design that avoids pooling entirely, directly aligning features from SR and HR images to preserve high-frequency and spatially sensitive distortion cues.

For SR FR-IQA, methods such as SIS, SFSN, and SRIF are primarily built upon hand-crafted features. While they incorporate insightful design concepts, these methods are constrained by shallow representations and limited capacity to exploit high-level semantic differences between SR outputs and HR references. This restricts their ability to fully model complex perceptual discrepancies in SR tasks. By contrast, our proposed PBAN fully leverages deep perceptual priors and attention-guided alignment, achieving consistent state-of-the-art results across QADS, CVIU, and Waterloo databases.
%Overall, the performance of PBAN validates its core design: bidirectional attention without pooling enables precise perception of enhanced distortions, while its perception-oriented architectures (i.e., GMDC and SubEC) integrate further fine-grained distortion sensitivity. These innovations enable PBAN to surpass prior SR IQA methods by a notable margin.

For visual comparison, we present some SR images along with the predicted quality scores by our model and another SR FR-IQA model in Fig. \ref{7}. Considering SFSN is one of the latest state-of-the-art SR FR-IQA metrics, and its code is easy to reproduce and validate, we take SFSN for comparison. Also, we provide the MOS of these images. Fig. \ref{7} shows that the SFSN model prefers
images with relatively low quality, such as Fig. \ref{7} (c), (d), which suffer apparent structural and textual distortion compared with their reference image. In contrast, the proposed method prefers Fig. \ref{7} (e), (f), which is consistent with human visual perception. Besides, the low score deviation between the proposed method and MOS indicates the superiority of our model on SR IQA.
\begin{table*}[t]
\footnotesize
  \centering
  \caption{Validity of Bi-Atten on QADS and CVIU databases. In order to separately demonstrate the role of Bi-Atten, all variants do not include GMDC and SubEC. HR → SR and SR → HR represent one-way “Key" transmission, while Bi-Atten has bidirectional “Key" transmission. “Key" transmission means the attention maps for one branch are calculated using the $K$ matrix from the other branch.}
    \begin{tabular}{c|c|cccc|cccc}
    \toprule
          &       & \multicolumn{4}{c|}{QADS}     & \multicolumn{4}{c}{CVIU} \\
    \midrule
    Types & Model & SRCC $\uparrow$  & KRCC $\uparrow$  & PLCC $\uparrow$  & RMSE $\downarrow$  & SRCC $\uparrow$  & KRCC $\uparrow$  & PLCC $\uparrow$  & RMSE $\downarrow$ \\
    \midrule
\multirow{4}[1]{*}{SR FR-IQA} & w/o Bi-Atten & 0.936 & 0.795 & 0.937 & 0.103 & 0.942 & 0.808 & 0.954 & 0.815 \\
          &  HR→SR & 0.957 & 0.840 & 0.958 & 0.079 & 0.970 & 0.853 & 0.971 & 0.706 \\
          & SR→HR & 0.955 & 0.834 & 0.957 & 0.082 & 0.960 & 0.829 & 0.960 & 0.763\\
          & with Bi-Atten & \textbf{0.981} & \textbf{0.895} & \textbf{0.982} & \textbf{0.055} & \textbf{0.972} & \textbf{0.862} & \textbf{0.976} & \textbf{0.515} \\
    \bottomrule
    \end{tabular}%
  \label{t5}%
\end{table*}%
\begin{table*}[t]
  \centering
  \scriptsize
  \caption{Validity of types of cross attention. General cross attention uses $K\&V$ homology, while the proposed Bi-Atten uses $Q\&V$ homology. Note that $Q\&K$ homology is self attention, and is shown in TABLE \ref{t5} as w/o Bi-Atten.}
    \begin{tabular}{c|cccc|cccc}
    \toprule
    Database & \multicolumn{4}{c|}{QADS}     & \multicolumn{4}{c}{CVIU} \\
    \midrule
    Same Source & SRCC $\uparrow$  & KRCC $\uparrow$ & PLCC $\uparrow$  & RMSE $\downarrow$  & SRCC $\uparrow$ & KRCC $\uparrow$  & PLCC $\uparrow$ & RMSE $\downarrow$\\
    \midrule
    K\&V  & 0.974 & 0.877 & 0.976 & \multicolumn{1}{c}{0.062} & 0.972 & 0.861 & 0.974 & 0.552 \\
    Q\&V  & \textbf{0.986} & \textbf{0.923} & \textbf{0.987} & \textbf{0.044} & \textbf{0.978} & \textbf{0.872} & \textbf{0.981} & \textbf{0.397} \\
    \bottomrule
    \end{tabular}
  \label{t6}%
\end{table*}%
\subsection{Ablation of Individual Proposed Components}
In this section, we conduct an ablation study to evaluate the effectiveness of each component in PBAN, as shown in TABLE \ref{t2}. The complete PBAN consistently achieves the best performance across all metrics, validating the overall design. Among the components, Bi-Atten contributes the most significant improvement, confirming its core role in aligning distortions from both the SR and HR perspectives. Both GMDC and SubEC also yield notable gains when used individually, and their integration with Bi-Atten further enhances performance. These results demonstrate that each component addresses a unique aspect of the perceptual quality modeling, and their joint use leads to strong complementary effects and the highest evaluation performance.
\subsection{Comparison of Model Efficiency}
We have conducted a computational efficiency comparison of seven advanced IQA approaches, including four general IQA models (Rows 1-4) and three SR IQA models (Rows 5-7), as shown in TABLE \ref{complexity}. The results reveal that the proposed PBAN demonstrates excellent computational efficiency and prediction performance. It has the second-lowest parameter count (2.222M) and low FLOPs (26.278G), while maintaining a fast inference speed (16.995 ms). Despite being lightweight, PBAN achieves the highest accuracy among all compared methods (SRCC = 0.986, PLCC = 0.987), significantly outperforming larger models. As for the limitations, PBAN is slightly slower than the most lightweight model (i.e., the DeepSRQ), due to the additional complexity introduced by the GMDC and SubEC. However, the accuracy gain justifies this small overhead.
\subsection{Validity of Bi-Atten}
\label{cross}
To validate the efficacy of Bi-Atten, we conduct a comparison on the attention interactive modes, as presented in TABLE \ref{t5}. Specifically, we test the following scenarios: without using Bi-Atten (i.e., applying the self-attention separately to the SR image and HR reference branches); transferring the “Key" information from the SR image branch to the HR reference branch while applying self-attention to the SR image branch (i.e., SR → HR); and transferring the “Key" information from the HR reference branch to the SR image branch while applying self-attention to the HR reference branch (i.e., HR → SR). Our experimental results reveal that the model without Bi-Atten exhibits the poorest performance on both databases, and even adding one-way “Key” information transmission could significantly improve the performance. Moreover, the performance improvement brought by Bi-Atten is particularly notable in KRCC and RMSE on both databases. Ultimately, the model using Bi-Atten achieves the best performance, indicating that our proposed Bi-Atten effectively provides the network's learning ability.

We further compare Bi-Atten (i.e., $Q\&V$ are homology) with cross attention (i.e., $K\&V$ are homology), and the results (TABLE \ref{t6}) show that Bi-Atten is superior in mining SR image distortion. Take the SR image branch as an example; this may be because using SR features as both query and value allows the model to project degraded information into the structural space of the reference, enabling more precise alignment and compensation for fine-grained distortions that are otherwise lost in cross attention; in contrast, when the query comes from the reference (i.e., $K\&V$ belong to SR space), it searches for correspondence in the degraded SR space, where incomplete reference information after projection hinders accurate alignment and leads to potential information loss.
\begin{table}[t]
\centering
\caption{The comparison between PBAN and spatial/channel attention on the CVIU database, with the best performance values highlighted in bold. The “spatial atten" and “channel atten" pool the channel and spatial dimensions to 1, respectively. The Flops is calculated with input tensor shape $3\times32\times32$.}
\resizebox{\linewidth}{!}{
\begin{tabular}{c|ccc|cccc}
\toprule
Model & Params/M & Flops/G & Speed/ms & SRCC $\uparrow$ & KRCC $\uparrow$ & PLCC $\uparrow$ & RMSE $\downarrow$ \\
\midrule 
Channel Atten & 2.222 & 0.181 & 11.284 & 0.963 & 0.836 & 0.964 & 0.716 \\
Spatial Atten & 2.222 & 0.181 & 11.099 & 0.913 & 0.751 & 0.917 & 1.062  \\
PBAN & 2.222 & 0.181 & 11.292 & \textbf{0.978} & \textbf{0.872} & \textbf{0.981} & \textbf{0.397} \\
\bottomrule
\end{tabular}}
\label{t10}%
\end{table}
\begin{table*}[t]
\centering
%\footnotesize
%\renewcommand{\arraystretch}{0.5}
\caption{Ablation study of GMDC on the CVIU databases with and without the SubEC. The variant “without GMDC" is to utilize standard convolution ($3\times 3$) to replace GMDC for experimentation. DCN refers to the standard deformable convolution, while the proposed GMDC uses grouped kernels denoted as $[K_1, K_2, \dots, K_N]$. The Flops are calculated with the input size $3\times32\times32$, and the inference speed is averaged over 100 runs. The best and second-best performances are highlighted in \textbf{bold} and \underline{underlined}, respectively.}
\resizebox{\linewidth}{!}{
\begin{tabular}{c|ccccccc|ccccccc}
\toprule
\multirow{2}{*}{Model} & \multicolumn{7}{c|}{Without SubEC} & \multicolumn{7}{c}{With SubEC} \\
\cmidrule{2-15} 
& Params/M & Flops/G & Speed/ms & SRCC $\uparrow$  & KRCC $\uparrow$  & PLCC $\uparrow$  & RMSE $\downarrow$ & Params/M & Flops/G & Speed/ms & SRCC $\uparrow$  & KRCC $\uparrow$  & PLCC $\uparrow$  & RMSE $\downarrow$ \\
\midrule
without GMDC & 1.387 & 0.045 & 3.933 & 0.972 & 0.862 & 0.976 & \textbf{0.515} & 1.407 & 0.062 & 7.946 & 0.976 & 0.871 & 0.977 & \underline{0.436} \\
with $3\times3$ DCN & 2.062 & 0.092 & 6.191 & 0.950 & 0.805 & 0.954 & 0.996 & 2.094 & 0.116 & 9.861 & 0.951 & 0.816 & 0.960 & 0.793 \\
with $5\times5$ DCN & 2.104 & 0.113 & 6.135 & 0.951 & 0.812 & 0.958 & 0.814 & 2.136 & 0.138 & 9.961 & 0.973 & 0.864 & 0.976 & 0.515 \\
with $7\times7$ DCN & 2.166 & 0.145 & 6.111 & 0.966 & 0.844 & 0.967 & 0.655 & 2.198 & 0.169 & 9.897 & 0.975 & 0.867 & 0.978 & 0.632 \\
with $9\times9$ DCN & 2.249 & 0.188 & 6.102 & 0.955 & 0.819 & 0.956 & 0.854 & 2.282 & 0.212 & 10.127 & 0.970 & 0.852 & 0.974 & 0.630 \\
with $[3,5]$ GMDC & 2.127 & 0.125 & 7.570 & 0.968 & 0.848 & 0.970 & 0.696 & 2.159 & 0.149 & 11.468 & 0.972 & 0.860 & 0.976 & 0.553 \\
with $[3,7]$ GMDC & 2.189 & 0.157 & 7.645 & \underline{0.974} & \underline{0.866} & \underline{0.977} & \underline{0.605} & 2.222 & 0.181 & 11.292 & \textbf{0.978} & \underline{0.872} & \textbf{0.981} & \textbf{0.397} \\
with $[3,9]$ GMDC & 2.273 & 0.199 & 7.486 & 0.971 & 0.854 & 0.972 & 0.669 & 2.305 & 0.224 & 11.639 & 0.973 & 0.862 & 0.975 & 0.630 \\
with $[3,5,7]$ GMDC & 2.254 & 0.190 & 8.719 & 0.972 & 0.857 & 0.972 & 0.642 & 2.286 & 0.214 & 13.144 & 0.975 & 0.871 & 0.978 & 0.521 \\
with $[5,7,9]$ GMDC & 2.442 & 0.285 & 8.941 & 0.967 & 0.845 & 0.968 & 0.718 & 2.474 & 0.310 & 13.011 & 0.974 & 0.864 & 0.976 & 0.581 \\
with $[3,5,7,9]$ GMDC & 2.465 & 0.297 & 10.554 & \textbf{0.975} & \textbf{0.868} & \textbf{0.978} & 0.632 & 2.497 & 0.322 & 14.418 & \underline{0.977} & \textbf{0.873} & \underline{0.980} & 0.501\\
\bottomrule
\end{tabular}}
\label{t3}%
\end{table*}%
\subsection{Validity on Replacing Bi-Atten with Spatial/Channel Attention}
\label{spatialchannel}
TABLE \ref{t10} compares PBAN with Spatial and Channel Attention in terms of complexity, inference speed, and performance metrics. All methods share identical model complexity (2.222M Params, 0.181G FLOPs); however, PBAN outperforms the others, achieving the highest SRCC (0.978), KRCC (0.872), PLCC (0.981), and the lowest RMSE (0.397). While PBAN's inference speed (11.292 ms) is slightly slower, the trade-off is justified by its substantial accuracy gains, highlighting its effectiveness and efficiency. This may be because SR images exhibit enhanced distortions, where spatial/channel attention—commonly used in common IQA with degraded distortions—may overlook critical information due to pooling operations.
\subsection{Validity of GMDC}
\subsubsection{The Individual Effectiveness}
As shown in TABLE \ref{t3}, the DCN does not consistently improve performance over the baseline “without GMDC", and in most cases, it even leads to performance degradation. For the $3\times3$ DCN, the receptive field is relatively small, limiting the space in which the deformable offsets can be effectively learned, often resulting in uninformative deformations. While moderate kernel sizes ($5\times5$ and $7\times7$) provide slight improvements due to a larger receptive field and more contextual information, performance drops again with the $9\times9$ kernel. This aligns with previous findings that standard DCN struggles to learn reliable offsets when the kernel size becomes larger~\cite{zhang2024ldconv}. In contrast, our proposed GMDC consistently delivers stable and significant improvements, especially with $[3,7]$ and $[3,5,7,9]$ settings. This is attributed to its grouped offset learning and multi-scale design, which enhances both the flexibility and robustness of spatial modeling. Note that there is no universally optimal combination; therefore, $[3,7]$ was empirically selected for its ability to consistently maximize performance while maintaining a reasonable computational overhead.
\begin{table}[t]
\centering
\caption{Validity for the point-wise convolution of GMDC on CVIU database. Note that the SubEC is not involved in these variants. The proposed GMDC uses grouped kernels denoted as $[K_1, K_2, \dots, K_N]$. The best performance is highlighted in bold.}
\resizebox{\linewidth}{!}{
\begin{tabular}{c|cccc|cccc}
\toprule
Model & \multicolumn{4}{c|}{w/o Point-wise Conv} & \multicolumn{4}{c}{with Point-wise Conv} \\
\midrule
GMDC Groups & SRCC $\uparrow$ & KRCC $\uparrow$ & PLCC $\uparrow$& RMSE $\downarrow$& SRCC $\uparrow$& KRCC $\uparrow$& PLCC $\uparrow$& RMSE $\downarrow$\\
\midrule
$[3,7]$ & 0.945 & 0.800 & 0.951 & 0.884 & 0.974 & 0.866 & 0.977 & 0.605 \\
$[3,5,7]$ & 0.967 & 0.850 & 0.968 & 0.713 & 0.972 & 0.857 & 0.972 & 0.642 \\
$[3,5,7,9]$ & 0.973 & 0.864 & 0.976 & \textbf{0.601} & \textbf{0.975} & \textbf{0.868} & \textbf{0.978} & 0.632\\
\bottomrule
\end{tabular}}
\label{GMDCpw}
\end{table}
\subsubsection{The Importance of Point-wise Convolution}
We have conducted an ablation study on the point-wise convolution in GMDC to demonstrate that the grouping strategy does indeed restrict cross-channel information interaction. And such a limitation of restricted cross-channel interaction is explicitly addressed by introducing a point-wise ($1\times1$) convolution after the grouped kernels. As shown in TABLE \ref{GMDCpw}, removing the point-wise convolution results in a noticeable performance drop across all evaluation metrics. For instance, in the configuration $[3,5,7]$, the PLCC decreases from 0.972 to 0.968, and the RMSE increases from 0.642 to 0.713. These results clearly confirm that the point-wise convolution plays a crucial role in compensating for the lack of inter-group communication, thereby enabling effective cross-channel information fusion following the grouped convolutions.
\subsubsection{The Impact of Channel Grouping}
As group convolution~\cite{resnext} is employed in GMDC, we conduct validation of the number of groups. Given that the number of channels for convolutional layers is set to a multiple of 16 and the group number needs to evenly divide the channel number, we test with groups of 2, 4, 8, and 16, respectively. The results are presented in TABLE \ref{t4}. It can be observed that GMDC with 2 groups yields optimal performance; however, as the group number doubles, there is a gradual decline in performance with more pronounced losses in terms of KRCC and RMSE. This suggests that there is a greater loss of inter-group information when the number of groups is super large (i.e., 16), even if the point-wise convolution is applied.
\begin{table}[t]
\tiny
  \centering
  \caption{Validity for the number of groups in group convolution of GMDC on the CVIU database. As multi-scale kernels are utilized, the minimum number of groups is 2. In experiments with different numbers of groups, half of the groups had kernel sizes set to 3, while the other half had kernel sizes set to 7.}
  \resizebox{\linewidth}{!}{
    \begin{tabular}{c|cccc}
    \toprule
    GMDC (Groups) & SRCC $\uparrow$ & KRCC $\uparrow$  & PLCC $\uparrow$  & RMSE $\downarrow$\\
    \midrule
    Groups 2     & \textbf{0.978} & \textbf{0.872} & \textbf{0.981} & \textbf{0.397} \\
   Groups 4     & 0.974 & 0.866 & 0.977 & 0.551 \\
   Groups 8     & 0.972 & 0.864 & 0.976 & 0.541 \\
   Groups 16    & 0.967 & 0.849 & 0.971 & 0.681 \\
    \bottomrule
    \end{tabular}}
  \label{t4}%
\end{table}%
\begin{table*}[t]
\footnotesize
  \centering
  \caption{Validity of SubEC on QADS database. Because it is proposed to further improve Bi-Atten, the performance is compared under different attention modes.}
    \begin{tabular}{c|c|cccc|cccc}
    \toprule
          &       & \multicolumn{4}{c|}{Without SubEC}     & \multicolumn{4}{c}{With SubEC} \\
    \midrule
    Types & Model & SRCC $\uparrow$ & KRCC $\uparrow$ & PLCC $\uparrow$ & RMSE $\downarrow$ & SRCC $\uparrow$ & KRCC $\uparrow$  & PLCC $\uparrow$ & RMSE $\downarrow$ \\
    \midrule
\multirow{4}[1]{*}{SR FR-IQA} & w/o Bi-Atten & 0.936 & 0.795 & 0.937 & 0.103 & 0.961 & 0.849 & 0.962 & 0.074 \\
          &  HR→SR & 0.957 & 0.840 & 0.958 & 0.079 & 0.982 & 0.911 & 0.984 & 0.050 \\
          & SR→HR & 0.955 & 0.834 & 0.957 & 0.082 & 0.983 & 0.915 & 0.985 & 0.048 \\
          & with Bi-Atten & \textbf{0.981} & \textbf{0.895} & \textbf{0.982} & \textbf{0.055} & \textbf{0.984} & \textbf{0.908} & \textbf{0.986} & \textbf{0.050} \\
    \bottomrule
    \end{tabular}%
  \label{t7}%
\end{table*}%
\begin{table}[t]
%\renewcommand{\arraystretch}{0.5}
%\scriptsize
  \centering
  \caption{Validity of the number of groups in group convolution in the Sub-Channel Weight branch of SubEC on the CVIU database. The best performance is highlighted in bold.}
  \resizebox{\linewidth}{!}{
    \begin{tabular}{c|ccc|cccc}
    \toprule
    \multicolumn{1}{c|}{} & \multicolumn{3}{c|}{Model Efficiency} & \multicolumn{4}{c}{Performance} \\
    \midrule
    Group Numbers & Params/M & Flops/G & Speed/ms & SRCC $\uparrow$ & KRCC $\uparrow$ & PLCC $\uparrow$ & RMSE $\downarrow$ \\
    \midrule
    Groups 2     & 2.222 & 0.181 & 11.292 &  \textbf{0.978} & \textbf{0.872} & \textbf{0.981} & \textbf{0.397} \\
   Groups 4     & 2.223 & 0.183 & 11.257 & 0.976 & 0.870 & 0.979 & 0.515 \\
   Groups 8     & 2.223 & 0.183 & 11.271 & 0.975 & 0.869 & 0.978 & 0.538 \\
   Groups 16    & 2.216 & 0.179 & 11.070 & 0.972 & 0.864 & 0.976 & 0.547 \\
   Groups 32 & 2.216 & 0.179 & 10.903 & 0.975 & 0.870 & 0.978 & 0.563 \\
    \bottomrule
    \end{tabular}}
  \label{subecg}%
\end{table}%
\subsection{Validity of SubEC}
\subsubsection{The Individual Effectiveness}
The SubEC is designed to effectively enhance the network’s ability to refine bi-directional attention for the perception of distorted information. To validate its effectiveness, experiments were conducted under different attention interaction modes, with the results presented in Table~\ref{t7}. Comparisons on the QADS database show that incorporating this component improves performance across all four attention interaction modes. In contrast, performance is notably worse when it is excluded, highlighting the substantial contribution of such design. These findings demonstrate not only its strong capability to capture fine-grained distorted information but also its ability to compensate for the limitations of one-way “Key” transmission.
\subsubsection{The Impact of Group Numbers}
As group convolution~\cite{resnext} is also employed in the Sub-Channel Weight branch, we conduct validation of the number of groups. As shown in TABLE \ref{subecg}, we tested with group counts of 2, 4, 8, 16, and 32, respectively. However, the results indicate that increasing the number of groups does not lead to significant or consistent performance improvements. Therefore, we set the number of groups to 2 for all other experiments.
\begin{table}[t]
  \centering
  \caption{The validity of the upsampling factor S in the Sub-Pixel Weight branch of SubEC. The best performance is highlighted in bold.}
  \resizebox{\linewidth}{!}{
    \begin{tabular}{c|ccc|cc|cc}
    \toprule
    \multicolumn{1}{c|}{}  &\multicolumn{3}{c|}{Model Efficiency} &\multicolumn{2}{c|}{CVIU} &\multicolumn{2}{c}{Waterloo} \\
    \midrule
    Upscale Factor S & Params/M & Flops/G & Speed/ms & SRCC $\uparrow$ & PLCC $\uparrow$ & SRCC $\uparrow$ & PLCC $\uparrow$ \\
    \midrule
    2 & 2.220 & 0.179 & 11.010 & \textbf{0.978} & \textbf{0.981} & 0.965& 0.979 \\
    3 & 2.223 & 0.183 & 11.174 & 0.977 & 0.980 & \textbf{0.970} & 0.980\\
    4 & 2.232 & 0.188 & 11.322 & 0.976 & 0.978 & 0.965 & 0.982\\
    5 & 2.241 & 0.195 & 11.205 & 0.978 & 0.981 & 0.965 & 0.980\\
    6 & 2.252 & 0.203 & 11.121 & 0.975 & 0.977 & 0.968 & \textbf{0.983} \\
    8 & 2.277 & 0.223 & 11.141 & 0.963 & 0.967 & 0.965 & 0.981\\
    \bottomrule
    \end{tabular}}
  \label{subecs}%
\end{table}%
\subsubsection{The Impact of magnification factor S}
Furthermore, we delve into the impact of the upsampling factor S for both channel and spatial dimensions on experimental outcomes. As shown in TABLE \ref{subecs}, we conducted an ablation study by varying S from 2 to 8; training and testing were individually conducted on the CVIU~\cite{cviu} and Waterloo~\cite{waterloo} databases. The results reveal that the upsampling factor S used in the Sub-Pixel Weight branch is not directly related to the SR scaling factors (e.g., the CVIU includes SR images generated with scaling factors of ×2, ×3, ×4, ×5, ×6, and ×8, while the Waterloo includes ×2, ×4, and ×8) used in the database. While model complexity obviously increases as S is up-scaled, the performance differences are generally small. In particular, S=2 consistently provides strong performance across all metrics, making it a reliable and empirically chosen setting. Therefore, S is not determined by the distribution of SR scales in the database, but rather chosen for its balanced effectiveness and efficiency in feature processing.
\begin{table}[t]
\centering
\caption{The impact of channel shuffle, dot-product, and upsampling methods in SubEC on CVIU database, with the best performance values highlighted in bold. The Flops is calculated with input tensor shape $3\times\ 32\times 32$.}
\resizebox{\linewidth}{!}{
\begin{tabular}{c|ccc|cccc}
\toprule
Model & Params/M & Flops/G & Speed/ms  & SRCC $\uparrow$ & KRCC $\uparrow$ & PLCC $\uparrow$ & RMSE $\downarrow$ \\
\midrule 
w/o Channel Shuffle & 2.220 & 0.179 & 10.981 & 0.970 & 0.853 & 0.971 & 0.654 \\
Element-wise Add & 2.220 & 0.179 & 10.690 & 0.963 & 0.838 & 0.966 & 0.699  \\
Transposed Conv & 2.194 & 0.160 & 10.905 & 0.975 & 0.871 & 0.979 & 0.528\\
Bilinear & 2.193 & 0.156 & 9.695 & 0.974 & 0.864 & 0.978 & 0.602 \\
Bicubic & 2.193 & 0.156 & 10.336 & 0.971 & 0.857 & 0.976 & 0.655 \\
PBAN (Sub-pixel Conv)  & 2.220 & 0.179 & 11.010 & \textbf{0.978} & \textbf{0.872} & \textbf{0.981} & \textbf{0.397} \\
\bottomrule
\end{tabular}}
\label{t11}
\end{table}
\subsubsection{The Impact of Channel Shuffle, Dot-product, and Upsampling Methods}
In this subsection, we further analyze the impact of channel shuffle and the replacement of dot-product with element-wise addition. As shown in TABLE \ref{t11}, removing the channel shuffle results in a noticeable performance drop, as the channel shuffle operation enhances the model's non-linear capability. Furthermore, replacing the dot product with element-wise addition also affects the results. Dot-product preserves the original distribution of input features, making its effects more controllable. In contrast, element-wise addition mixes the effects of channels and spatial features, changing the feature distribution and potentially compromising performance. 

Moreover, we evaluate the impact of different upsampling strategies by comparing PBAN (Sub-pixel Convolution) with three alternatives: transposed convolution, Bilinear interpolation, and Bicubic interpolation. The results demonstrate that the choice of upsampling method significantly affects performance. Specifically, PBAN and transposed convolution achieve the best and the second-best SRCC/PLCC scores (i.e., 0.978/0.981 and 0.975/0.979), respectively. While interpolation-based methods (Bilinear and Bicubic) yield slightly inferior results. This can be attributed to the fact that adopting a learning-based upsampling approach like transposed convolution proves to be both effective and well-suited for this task. Besides, Sub-pixel convolution, originally designed for SR image reconstruction tasks, aligns best with the objectives of SR IQA.
\subsection{Cross-dataset Evaluation}
We conduct cross-dataset tests to validate the generalization performance of our proposed PBAN and five deep learning-based comparison methods. In practice, we train algorithms on one database and test them on the other. We present SRCC, KRCC, and PLCC results of the cross-dataset tests in TABLE \ref{t8}, where the best and second-best results are labeled in \textcolor{red}{red} and \textcolor{blue}{blue}, respectively. TABLE \ref{t8} reveals that PBAN achieves desirable generalization performance on benchmark SR-IQA databases, even when training on a small database (QADS) and testing on the relatively larger database (CVIU). 
\begin{table}[t]
\small
  \centering
  \caption{The performance for cross-dataset evaluation, where the best and suboptimal performance values are in \textcolor{red}{Red} and \textcolor{blue}{Blue}, respectively.}
  \resizebox{\linewidth}{!}{
    \begin{tabular}{c|ccc|ccc}
    \toprule
    Train & \multicolumn{3}{c|}{CVIU} & \multicolumn{3}{c}{QADS} \\
    \midrule
    Test  & \multicolumn{3}{c|}{QADS} & \multicolumn{3}{c}{CVIU} \\
    \midrule
    Model & SRCC $\uparrow$ & KRCC $\uparrow$ & PLCC $\uparrow$ & SRCC $\uparrow$ & KRCC $\uparrow$ & PLCC $\uparrow$ \\
    \midrule
    DeepSRQ & 0.723  &\makebox[1em][c]{\textemdash}     & 0.749  & 0.648  &\makebox[1em][c]{\textemdash}     &\makebox[1em][c]{\textemdash} \\
    HLSRIQA & \textcolor{blue}{0.786}  &\makebox[1em][c]{\textemdash}     & \textcolor{blue}{0.773}  & $\;$\textcolor{red}{0.785 } &\makebox[1em][c]{\textemdash}     & \textcolor{blue}{0.776}  \\
    EK-SR-IQA & 0.645  &\makebox[1em][c]{\textemdash}     &\makebox[1em][c]{\textemdash}     & 0.682  &\makebox[1em][c]{\textemdash}     &\makebox[1em][c]{\textemdash} \\
    JCSAN & 0.702  & 0.530  & 0.710  & 0.631  & 0.463  & 0.729  \\
    TADSRNet & 0.739  & 0.547  & 0.736  & 0.657  & 0.489  & 0.729  \\
    PBAN & \textcolor[rgb]{ 1,  0,  0}{$\;$0.890 } & $\;$\textcolor{red}{0.732 } & $\;$\textcolor{red}{0.892 } & \textcolor{blue}{0.730}  & $\;$\textcolor{red}{0.550 } & $\;$\textcolor{red}{0.794 } \\
    \bottomrule
    \end{tabular}}
  \label{t8}%
\end{table}%
\begin{table}[t]
\small
  \centering
  \caption{The performance of common NR-IQA evaluation, with the best and suboptimal performance values highlighted in \textcolor{red}{red} and \textcolor{blue}{blue}, respectively. The Flops is calculated with input tensor shape $3\times\ 512\times 512$.}
  \resizebox{\linewidth}{!}{
\begin{tabular}{c|ccc|cc|cr}
\midrule \multicolumn{1}{c|}{} & \multicolumn{3}{c|}{Model Efficiency} & \multicolumn{2}{c|}{ CLIVE } & \multicolumn{2}{c}{ KonIQ-10K } \\
\midrule Model & Params/M & Flops/G & Speed/Ms & PLCC $\uparrow$ & SRCC $\uparrow$ & PLCC $\uparrow$ & SRCC $\uparrow$\\
\midrule WaDIQaM~\cite{bosse} & 6.287 & 30.479 & 8.170 & 0.671 & 0.682 & 0.807 & 0.804 \\
DBCNN~\cite{DBCNN} & 15.311 & 86.219 & 23.688 & 0.869 & 0.869 & 0.884 & 0.875 \\
CNNIQA~\cite{CNNIQA} & 0.729 & 1.883 & 71.210 & 0.450 & 0.465 & 0.584 & 0.572 \\
HyperIQA~\cite{hyperiqa} & 27.375 & 107.831 & 29.980 & 0.882 & 0.859 & 0.917 & 0.906 \\
MetaIQA~\cite{metaiqa} &\makebox[1em][c]{\textemdash} &\makebox[1em][c]{\textemdash} &\makebox[1em][c]{\textemdash} & 0.802 & 0.835 & 0.856 & 0.887 \\
MUSIQ~\cite{ke2021musiq} & 27.126 & 17.646 & 14.910 & 0.832 & 0.793 & 0.928 & \textcolor{blue}{0.916} \\
CLIP-IQA~\cite{clipiqa} &\makebox[1em][c]{\textemdash} & 61.065 & 21.100 & 0.831 & 0.805 & 0.845 & 0.803 \\
TReS~\cite{tres} & 152.452 & 500.040 & 251.070 & 0.877 & 0.846 & 0.928 & \textcolor{blue}{0.916} \\
TOPIQ~\cite{chen2024topiq} & 36.039 & 50.138 & 22.140 & \textcolor{blue}{0.884} & \textcolor{blue}{0.870} & \textcolor{red}{0.939} & \textcolor{red}{0.926} \\
PBAN-NR & 1.598 & 46.632 & 18.219 & \textcolor{red}{0.899} & \textcolor{red}{0.886} & 0.924 & \textcolor{blue}{0.916} \\
\midrule
\end{tabular}}
\label{t9}%
\end{table}%
\subsection{Validity on Common IQA}
TABLE~\ref{t9} compares the performance of advanced common IQA methods on the CLIVE~\cite{livec} and KonIQ-10K~\cite{hosu2020koniq} databases, following the protocol of TOPIQ~\cite{chen2024topiq}, evaluated by parameters (Params), computational complexity (Flops), and correlation metrics (PLCC and SRCC). We replaced the Bi-Atten with self-attention to form the PBA-NR block. The PBAN-NR utilizes a single branch (four stacking PBA-NR blocks) to take distorted images as input and predict quality without reference information. The results demonstrate that the PBAN-NR model, with the smallest parameter size (1.598M) and relatively low computational cost (46.632G), achieves the best or second-best performance across both databases. Furthermore, compared to models with significantly larger parameter sizes, such as HyperIQA and TReS, PBAN-NR strikes a better balance between efficiency and performance.

\section{Conclusion}
In this paper, we propose a deep learning-based SR FR-IQA method called PBAN, following the inspiration from the attention characteristic, hierarchy, and spatial frequency sensitivity of the HVS. Motivated by the generation and evaluation processes of SR distortion, we introduce Bi-Atten. This creates a novel attention mechanism for paying visual attention to distortion in an iterative manner. To provide the perception of distortion by utilizing the adaptive dense spatial prediction advantages of deformable convolution, we propose GMDC that balances multi-scale modeling and computational complexity. Inspired by sub-pixel methods in the field of camera imaging, we design SubEC to further refine the perception by focusing on mining finer distorted information. Experimental results demonstrate that our proposed PBAN effectively provides visual perception to SR distortion and surpasses existing state-of-the-art quality assessment methods. 

In the future, we plan to construct a large-scale SR quality database containing the latest SR models (e.g., GAN-based methods and diffusion model-based methods) to enhance the generalization of SR IQA methods. Moreover, we will design specific SR quality metrics for videos and develop quality-driven SR algorithms.
%\begin{spacing}{0.9}
\small
\bibliographystyle{IEEEtran}
\bibliography{bibfile}

\begin{thebibliography}{10}
\providecommand{\url}[1]{#1}
\csname url@samestyle\endcsname
\providecommand{\newblock}{\relax}
\providecommand{\bibinfo}[2]{#2}
\providecommand{\BIBentrySTDinterwordspacing}{\spaceskip=0pt\relax}
\providecommand{\BIBentryALTinterwordstretchfactor}{4}
\providecommand{\BIBentryALTinterwordspacing}{\spaceskip=\fontdimen2\font plus
\BIBentryALTinterwordstretchfactor\fontdimen3\font minus \fontdimen4\font\relax}
\providecommand{\BIBforeignlanguage}[2]{{%
\expandafter\ifx\csname l@#1\endcsname\relax
\typeout{** WARNING: IEEEtran.bst: No hyphenation pattern has been}%
\typeout{** loaded for the language `#1'. Using the pattern for}%
\typeout{** the default language instead.}%
\else
\language=\csname l@#1\endcsname
\fi
#2}}
\providecommand{\BIBdecl}{\relax}
\BIBdecl

\bibitem{peng2024efficient}
L.~Peng, Y.~Cao, R.~Pei, W.~Li, J.~Guo, X.~Fu, Y.~Wang, and Z.-J. Zha, ``Efficient {Real}-world {Image} {Super}-{Resolution} {Via} {Adaptive} {Directional} {Gradient} {Convolution},'' \emph{arXiv preprint arXiv:2405.07023}, 2024.

\bibitem{lee2025emulating}
D.~Lee, S.~Yun, and Y.~Ro, ``Emulating {Self}-attention with {Convolution} for {Efficient} {Image} {Super}-{Resolution},'' \emph{arXiv preprint arXiv:2503.06671}, 2025.

\bibitem{zhang2019ranksrgan}
W.~Zhang, Y.~Liu, C.~Dong, and Y.~Qiao, ``{RankSRGAN}: {Generative} {Adversarial} {Networks} {With} {Ranker} for {Image} {Super}-{Resolution},'' in \emph{Proceedings of the IEEE/CVF International Conference on Computer Vision}, 2019, pp. 3096--3105.

\bibitem{li2024sed}
B.~Li, X.~Li, H.~Zhu, Y.~Jin, R.~Feng, Z.~Zhang, and Z.~Chen, ``{SeD}: {Semantic}-{Aware} {Discriminator} for {Image} {Super}-{Resolution},'' in \emph{Proceedings of the IEEE/CVF Conference on Computer Vision and Pattern Recognition}, 2024, pp. 25\,784--25\,795.

\bibitem{wang2024camixersr}
Y.~Wang, Y.~Liu, S.~Zhao, J.~Li, and L.~Zhang, ``{CAMixerSR}: {Only} {Details} {Need} {More} {“Attention"},'' in \emph{Proceedings of the IEEE/CVF Conference on Computer Vision and Pattern Recognition}, 2024, pp. 25\,837--25\,846.

\bibitem{park2025efficient}
K.~Park, J.~W. Soh, and N.~I. Cho, ``Efficient {Attention}-{Sharing} {Information} {Distillation} {Transformer} for {Lightweight} {Single} {Image} {Super}-{Resolution},'' \emph{arXiv preprint arXiv:2501.15774}, 2025.

\bibitem{tian2024image}
Y.~Tian, H.~Chen, C.~Xu, and Y.~Wang, ``Image {Processing} {GNN}: {Breaking} {Rigidity} in {Super}-{Resolution},'' in \emph{Proceedings of the IEEE/CVF Conference on Computer Vision and Pattern Recognition}, 2024, pp. 24\,108--24\,117.

\bibitem{AttenSR}
Y.~Hu, J.~Li, Y.~Huang, and X.~Gao, ``Channel-{Wise} and {Spatial} {Feature} {Modulation} {Network} for {Single} {Image} {Super}-{Resolution},'' \emph{IEEE Transactions on Circuits and Systems for Video Technology}, vol.~30, no.~11, pp. 3911--3927, 2020.

\bibitem{yang2025diffusion}
J.~Yang, T.~Dai, Y.~Zhu, N.~Li, J.~Li, and S.-T. Xia, ``Diffusion {Prior} {Interpolation} for {Flexibility} {Real}-{World} {Face} {Super}-{Resolution},'' in \emph{Proceedings of the AAAI Conference on Artificial Intelligence}, vol.~39, no.~9, 2025, pp. 9211--9219.

\bibitem{li2025difiisr}
X.~Li, Z.~Wang, Y.~Zou, Z.~Chen, J.~Ma, Z.~Jiang, L.~Ma, and J.~Liu, ``{DifIISR}: {A} {Diffusion} {Model} with {Gradient} {Guidance} for {Infrared} {Image} {Super}-{Resolution},'' in \emph{Proceedings of the IEEE/CVF Computer Vision and Pattern Recognition Conference}, 2025, pp. 7534--7544.

\bibitem{moser2025dynamic}
B.~B. Moser, S.~Frolov, F.~Raue, S.~Palacio, and A.~Dengel, ``Dynamic {Attention}-{Guided} {Diffusion} for {Image} {Super}-{Resolution},'' in \emph{2025 IEEE/CVF Winter Conference on Applications of Computer Vision (WACV)}, 2025, pp. 451--460.

\bibitem{wu2024seesr}
R.~Wu, T.~Yang, L.~Sun, Z.~Zhang, S.~Li, and L.~Zhang, ``{SeeSR}: {Towards} {Semantics}-{Aware} {Real}-{World} {Image} {Super}-{Resolution},'' in \emph{Proceedings of the IEEE/CVF Computer Vision and Pattern Recognition Conference}, 2024, pp. 25\,456--25\,467.

\bibitem{yu2024scaling}
F.~Yu, J.~Gu, Z.~Li, J.~Hu, X.~Kong, X.~Wang, J.~He, Y.~Qiao, and C.~Dong, ``Scaling {Up} to {Excellence}: {Practicing} {Model} {Scaling} for {Photo}-{Realistic} {Image} {Restoration} {In} the {Wild},'' in \emph{Proceedings of the IEEE/CVF Conference on Computer Vision and Pattern Recognition}, 2024, pp. 25\,669--25\,680.

\bibitem{ssim}
Z.~Wang, A.~Bovik, H.~Sheikh, and E.~Simoncelli, ``Image quality assessment: from error visibility to structural similarity,'' \emph{IEEE Transactions on Image Processing}, vol.~13, no.~4, pp. 600--612, 2004.

\bibitem{cwssim}
M.~P. Sampat, Z.~Wang, S.~Gupta, A.~C. Bovik, and M.~K. Markey, ``Complex {Wavelet} {Structural} {Similarity}: {A} {New} {Image} {Similarity} {Index},'' \emph{IEEE Transactions on Image Processing}, vol.~18, no.~11, pp. 2385--2401, 2009.

\bibitem{msssim}
Z.~Wang, E.~P. Simoncelli, and A.~C. Bovik, ``Multiscale structural similarity for image quality assessment,'' in \emph{The Thrity-Seventh Asilomar Conference on Signals, Systems \& Computers, 2003}, vol.~2.\hskip 1em plus 0.5em minus 0.4em\relax IEEE, 2003, pp. 1398--1402.

\bibitem{bosse}
S.~Bosse, D.~Maniry, K.-R. Müller, T.~Wiegand, and W.~Samek, ``Deep {Neural} {Networks} for {No}-{Reference} and {Full}-{Reference} {Image} {Quality} {Assessment},'' \emph{IEEE Transactions on Image Processing}, vol.~27, no.~1, pp. 206--219, 2018.

\bibitem{DBCNN}
W.~Zhang, K.~Ma, J.~Yan, D.~Deng, and Z.~Wang, ``Blind {Image} {Quality} {Assessment} {Using} a {Deep} {Bilinear} {Convolutional} {Neural} {Network},'' \emph{IEEE Transactions on Circuits and Systems for Video Technology}, vol.~30, no.~1, pp. 36--47, 2020.

\bibitem{qiup2}
Q.~Jiang, F.~Liu, Z.~Wang, S.~Wang, and W.~Lin, ``Rethinking and {Conceptualizing} {Just} {Noticeable} {Difference} {Estimation} by {Residual} {Learning},'' \emph{IEEE Transactions on Circuits and Systems for Video Technology}, vol.~34, no.~10, pp. 9515--9527, 2024.

\bibitem{metaiqa}
H.~Zhu, L.~Li, J.~Wu, W.~Dong, and G.~Shi, ``{MetaIQA}: {Deep} {Meta}-{Learning} for {No}-{Reference} {Image} {Quality} {Assessment},'' in \emph{Proceedings of the IEEE/CVF Conference on Computer Vision and Pattern Recognition}, 2020, pp. 14\,143--14\,152.

\bibitem{qiup3}
Q.~Jiang, Y.~Gu, Z.~Wu, C.~Li, H.~Xiong, F.~Shao, and Z.~Wang, ``Deep {Underwater} {Image} {Quality} {Assessment} {With} {Explicit} {Degradation} {Awareness} {Embedding},'' \emph{IEEE Transactions on Image Processing}, vol.~34, pp. 1297--1310, 2025.

\bibitem{hyperiqa}
S.~Su, Q.~Yan, Y.~Zhu, C.~Zhang, X.~Ge, J.~Sun, and Y.~Zhang, ``Blindly {Assess} {Image} {Quality} in the {Wild} {Guided} by a {Self}-{Adaptive} {Hyper} {Network},'' in \emph{Proceedings of the IEEE/CVF Conference on Computer Vision and Pattern Recognition}, 2020, pp. 3667--3676.

\bibitem{LIQA}
J.~Liu, W.~Zhou, X.~Li, J.~Xu, and Z.~Chen, ``{LIQA}: {Lifelong} {Blind} {Image} {Quality} {Assessment},'' \emph{IEEE Transactions on Multimedia}, vol.~25, pp. 5358--5373, 2023.

\bibitem{gu2020image}
J.~Gu, H.~Cai, H.~Chen, X.~Ye, J.~Ren, and C.~Dong, ``Image {Quality} {Assessment} for {Perceptual} {Image} {Restoration}: {A} {New} {Dataset}, {Benchmark} and {Metric},'' \emph{arXiv preprint arXiv:2011.15002}, 2020.

\bibitem{9937738}
Z.~Zhang, W.~Sun, X.~Min, W.~Zhu, T.~Wang, W.~Lu, and G.~Zhai, ``A {No}-{Reference} {Deep} {Learning} {Quality} {Assessment} {Method} for {Super}-{Resolution} {Images} {Based} on {Frequency} {Maps},'' in \emph{2022 IEEE International Symposium on Circuits and Systems (ISCAS)}, 2022, pp. 3170--3174.

\bibitem{sis}
F.~Zhou, R.~Yao, B.~Liu, and G.~Qiu, ``Visual {Quality} {Assessment} for {Super}-{Resolved} {Images}: {Database} and {Method},'' \emph{IEEE Transactions on Image Processing}, vol.~28, no.~7, pp. 3528--3541, 2019.

\bibitem{sfsn}
W.~Zhou, Z.~Wang, and Z.~Chen, ``Image {Super}-{Resolution} {Quality} {Assessment}: {Structural} {Fidelity} {Versus} {Statistical} {Naturalness},'' in \emph{2021 13th International conference on quality of multimedia experience (QoMEX)}.\hskip 1em plus 0.5em minus 0.4em\relax IEEE, 2021, pp. 61--64.

\bibitem{srif}
W.~Zhou and Z.~Wang, ``Quality {Assessment} of {Image} {Super}-{Resolution}: {Balancing} {Deterministic} and {Statistical} {Fidelity}@,'' in \emph{Proceedings of the 30th ACM International Conference on Multimedia}, 2022, pp. 934--942.

\bibitem{deepsqr}
W.~Zhou, Q.~Jiang, Y.~Wang, Z.~Chen, and W.~Li, ``Blind quality assessment for image superresolution using deep two-stream convolutional networks,'' \emph{Information Sciences}, vol. 528, pp. 205--218, 2020.

\bibitem{HLSRIQA}
Z.~Zhang, W.~Sun, X.~Min, W.~Zhu, T.~Wang, W.~Lu, and G.~Zhai, ``A {No}-{Reference} {Deep} {Learning} {Quality} {Assessment} {Method} for {Super}-{Resolution} {Images} {Based} on {Frequency} {Maps},'' in \emph{IEEE International Symposium on Circuits and Systems (ISCAS)}, 2022, pp. 3170--3174.

\bibitem{disq}
T.~Zhao, Y.~Lin, Y.~Xu, W.~Chen, and Z.~Wang, ``Learning-{Based} {Quality} {Assessment} for {Image} {Super}-{Resolution},'' \emph{IEEE Transactions on Multimedia}, vol.~24, pp. 3570--3581, 2021.

\bibitem{eksriqa}
H.~Zhang, S.~Su, Y.~Zhu, J.~Sun, and Y.~Zhang, ``Boosting {No}-{Reference} {Super}-{Resolution} {Image} {Quality} {Assessment} with {Knowledge} {Distillation} and {Extension},'' in \emph{IEEE International Conference on Acoustics, Speech and Signal Processing (ICASSP)}, 2023, pp. 1--5.

\bibitem{jcsan}
T.~Zhang, K.~Zhang, C.~Xiao, Z.~Xiong, and J.~Lu, ``Joint channel-spatial attention network for super-resolution image quality assessment,'' \emph{Applied Intelligence}, vol.~52, no.~15, pp. 17\,118--17\,132, 2022.

\bibitem{tadsrnet}
X.~Quan, K.~Zhang, H.~Li, D.~Fan, Y.~Hu, and J.~Chen, ``{TADSRNet}: {A} triple-attention dual-scale residual network for super-resolution image quality assessment,'' \emph{Applied Intelligence}, vol.~53, no.~22, pp. 26\,708--26\,724, 2023.

\bibitem{dai_deformable_2017}
J.~Dai, H.~Qi, Y.~Xiong, Y.~Li, G.~Zhang, H.~Hu, and Y.~Wei, ``Deformable {Convolutional} {Networks},'' in \emph{Proceedings of the IEEE/CVF International Conference on Computer Vision}, 2017, pp. 764--773.

\bibitem{zhu_deformable_2019}
X.~Zhu, H.~Hu, S.~Lin, and J.~Dai, ``Deformable {ConvNets} {V2}: {More} {Deformable}, {Better} {Results},'' in \emph{Proceedings of the IEEE/CVF Conference on Computer Vision and Pattern Recognition}, 2019, pp. 9308--9316.

\bibitem{spatialofHVS}
W.~Kirsch and W.~Kunde, ``Human perception of spatial frequency varies with stimulus orientation and location in the visual field,'' \emph{Scientific Reports}, vol.~13, no.~1, p. 17656, 2023.

\bibitem{radn}
S.~Shi, Q.~Bai, M.~Cao, W.~Xia, J.~Wang, Y.~Chen, and Y.~Yang, ``Region-{Adaptive} {Deformable} {Network} for {Image} {Quality} {Assessment},'' in \emph{Proceedings of the IEEE/CVF Conference on Computer Vision and Pattern Recognition}, 2021, pp. 324--333.

\bibitem{dqm-iqa}
Z.~Shi, Z.~Wang, F.~Kong, R.~Li, and T.~Luo, ``Dual-quality map based no reference image quality assessment using deformable convolution,'' \emph{Digital Signal Processing}, vol. 123, p. 103398, 2022.

\bibitem{hierarchy}
S.~Hochstein and M.~Ahissar, ``View from the {Top}: {Hierarchies} and {Reverse} {Hierarchies} in the {Visual} {System},'' \emph{Neuron}, vol.~36, no.~5, pp. 791--804, 2002.

\bibitem{espcn}
W.~Shi, J.~Caballero, F.~Huszar, J.~Totz, A.~P. Aitken, R.~Bishop, D.~Rueckert, and Z.~Wang, ``Real-{Time} {Single} {Image} and {Video} {Super}-{Resolution} {Using} an {Efficient} {Sub}-{Pixel} {Convolutional} {Neural} {Network},'' in \emph{Proceedings of the IEEE/CVF Conference on Computer Vision and Pattern Recognition}, 2016, pp. 1874--1883.

\bibitem{qssim}
A.~Kolaman and O.~Yadid-Pecht, ``Quaternion {Structural} {Similarity}: {A} {New} {Quality} {Index} for {Color} {Images},'' \emph{IEEE Transactions on Image Processing}, vol.~21, no.~4, pp. 1526--1536, 2012.

\bibitem{gmsd}
W.~Xue, L.~Zhang, X.~Mou, and A.~C. Bovik, ``Gradient {Magnitude} {Similarity} {Deviation}: {A} {Highly} {Efficient} {Perceptual} {Image} {Quality} {Index},'' \emph{IEEE Transactions on Image Processing}, vol.~23, no.~2, pp. 684--695, 2014.

\bibitem{MGCN}
C.~Huang, T.~Jiang, and M.~Jiang, ``Encoding {Distortions} for {Multi}-task {Full}-{Reference} {Image} {Quality} {Assessment},'' in \emph{2019 IEEE International Conference on Multimedia and Expo (ICME)}.\hskip 1em plus 0.5em minus 0.4em\relax IEEE, 2019, pp. 1864--1869.

\bibitem{spsim}
W.~Sun, Q.~Liao, J.-H. Xue, and F.~Zhou, ``{SPSIM}: {A} {Superpixel}-{Based} {Similarity} {Index} for {Full}-{Reference} {Image} {Quality} {Assessment},'' \emph{IEEE Transactions on Image Processing}, vol.~27, no.~9, pp. 4232--4244, 2018.

\bibitem{Zhang_2018_CVPR}
R.~Zhang, P.~Isola, A.~A. Efros, E.~Shechtman, and O.~Wang, ``The {Unreasonable} {Effectiveness} of {Deep} {Features} as a {Perceptual} {Metric},'' in \emph{Proceedings of the IEEE Conference on Computer Vision and Pattern Recognition}, June 2018.

\bibitem{fgiqa}
Z.~Zhang, W.~Sun, X.~Min, T.~Wang, W.~Lu, and G.~Zhai, ``A {Full}-{Reference} {Quality} {Assessment} {Metric} for {Fine}-{Grained} {Compressed} {Images},'' in \emph{International Conference on Visual Communications and Image Processing}, 2021, pp. 1--4.

\bibitem{mfan}
H.~Li and X.~Zhang, ``{MFAN}: {A} {Multi}-{Projection} {Fusion} {Attention} {Network} for {No}-{Reference} and {Full}-{Reference} {Panoramic} {Image} {Quality} {Assessment},'' \emph{IEEE Signal Processing Letters}, vol.~30, pp. 1207--1211, 2023.

\bibitem{doss}
X.~Liao, X.~Wei, M.~Zhou, and S.~Kwong, ``Full-{Reference} {Image} {Quality} {Assessment}: {Addressing} {Content} {Misalignment} {Issue} by {Comparing} {Order} {Statistics} of {Deep} {Features},'' \emph{IEEE Transactions on Broadcasting}, vol.~70, no.~1, pp. 305--315, 2024.

\bibitem{grids}
W.~Shen, M.~Zhou, J.~Luo, Z.~Li, and S.~Kwong, ``Graph-{Represented} {Distribution} {Similarity} {Index} for {Full}-{Reference} {Image} {Quality} {Assessment},'' \emph{IEEE Transactions on Image Processing}, vol.~33, pp. 3075--3089, 2024.

\bibitem{dmm}
B.~Chen, H.~Zhu, L.~Zhu, S.~Wang, J.~Pan, and S.~Wang, ``Debiased {Mapping} for {Full}-{Reference} {Image} {Quality} {Assessment},'' \emph{IEEE Transactions on Multimedia}, vol.~27, pp. 2638--2649, 2025.

\bibitem{kltsrqa}
Q.~Jiang, Z.~Liu, K.~Gu, F.~Shao, X.~Zhang, H.~Liu, and W.~Lin, ``Single {Image} {Super}-{Resolution} {Quality} {Assessment}: {A} {Real}-{World} {Dataset}, {Subjective} {Studies}, and an {Objective} {Metric},'' \emph{IEEE Transactions on Image Processing}, vol.~31, pp. 2279--2294, 2022.

\bibitem{C2MT}
H.~Li, K.~Zhang, Z.~Niu, and H.~Shi, ``C$^{2}${MT}: {A} {Credible} and {Class}-{Aware} {Multi}-{Task} {Transformer} for {SR-IQA},'' \emph{IEEE Signal Processing Letters}, vol.~29, pp. 2662--2666, 2022.

\bibitem{deformsr}
G.~Li, L.~Qiu, H.~Zhang, F.~Xie, and Z.~Jiang, ``Multi-{Frame} {Super}-{Resolution} {With} {Raw} {Images} {Via} {Modified} {Deformable} {Convolution},'' in \emph{2022 IEEE International Conference on Acoustics, Speech and Signal Processing (ICASSP)}.\hskip 1em plus 0.5em minus 0.4em\relax IEEE, 2022, pp. 2155--2159.

\bibitem{edvr}
X.~Wang, K.~C. Chan, K.~Yu, C.~Dong, and C.~C. Loy, ``{EDVR}: {Video} {Restoration} {With} {Enhanced} {Deformable} {Convolutional} {Networks},'' in \emph{Proceedings of the IEEE/CVF Conference on Computer Vision and Pattern Recognition}, 2019, pp. 1954--1963.

\bibitem{subpixelreconstruction}
L.~Yu, X.~Zhang, and Y.~Chu, ``Super-{Resolution} {Reconstruction} {Algorithm} for {Infrared} {Image} with {Double} {Regular} {Items} {Based} on {Sub}-{Pixel} {Convolution},'' \emph{Applied Sciences}, vol.~10, no.~3, 2020.

\bibitem{subpixelms}
J.~Qu, J.~Yin, Y.~Jiang, W.~Huang, and Q.~Chen, ``A sub-pixel convolution-based improved bidirectional feature pyramid network for pansharpening,'' \emph{Remote Sensing Letters}, vol.~14, no.~1, pp. 91--101, 2023.

\bibitem{crossatten}
C.-F.~R. Chen, Q.~Fan, and R.~Panda, ``{CrossViT}: {Cross}-{Attention} {Multi}-{Scale} {Vision} {Transformer} for {Image} {Classification},'' in \emph{Proceedings of the IEEE/CVF International Conference on Computer Vision}, 2021, pp. 347--356.

\bibitem{resnext}
S.~Xie, R.~Girshick, P.~Dollár, Z.~Tu, and K.~He, ``Aggregated {Residual} {Transformations} for {Deep} {Neural} {Networks},'' in \emph{Proceedings of the IEEE/CVF Conference on Computer Vision and Pattern Recognition}, 2017, pp. 5987--5995.

\bibitem{shufflenet}
X.~Zhang, X.~Zhou, M.~Lin, and J.~Sun, ``{ShuffleNet}: {An} {Extremely} {Efficient} {Convolutional} {Neural} {Network} for {Mobile} {Devices},'' in \emph{Proceedings of the IEEE/CVF Conference on Computer Vision and Pattern Recognition}, 2018, pp. 6848--6856.

\bibitem{resnet}
K.~He, X.~Zhang, S.~Ren, and J.~Sun, ``Deep {Residual} {Learning} for {Image} {Recognition},'' in \emph{Proceedings of the IEEE/CVF Conference on Computer Vision and Pattern Recognition}, 2016.

\bibitem{cviu}
C.~Ma, C.-Y. Yang, X.~Yang, and M.-H. Yang, ``Learning a no-reference quality metric for single-image super-resolution,'' \emph{Computer Vision and Image Understanding}, vol. 158, pp. 1--16, 2017.

\bibitem{waterloo}
H.~Yeganeh, M.~Rostami, and Z.~Wang, ``Objective {Quality} {Assessment} of {Interpolated} {Natural} {Images},'' \emph{IEEE Transactions on Image Processing}, vol.~24, no.~11, pp. 4651--4663, 2015.

\bibitem{LPIPS}
R.~Zhang, P.~Isola, A.~A. Efros, E.~Shechtman, and O.~Wang, ``The {Unreasonable} {Effectiveness} of {Deep} {Features} as a {Perceptual} {Metric},'' in \emph{Proceedings of the IEEE/CVF Conference on Computer Vision and Pattern Recognition}, 2018, pp. 586--595.

\bibitem{niqe}
A.~Mittal, R.~Soundararajan, and A.~C. Bovik, ``Making a “{Completely} {Blind}” {Image} {Quality} {Analyzer},'' \emph{IEEE Signal Processing Letters}, vol.~20, no.~3, pp. 209--212, 2013.

\bibitem{lpsi}
Q.~Wu, Z.~Wang, and H.~Li, ``A highly efficient method for blind image quality assessment,'' in \emph{2015 IEEE International Conference on Image Processing (ICIP)}, 2015, pp. 339--343.

\bibitem{clipiqa}
J.~Wang, K.~C. Chan, and C.~C. Loy, ``Exploring {CLIP} for {Assessing} the {Look} and {Feel} of {Images},'' in \emph{Proceedings of the AAAI Conference on Artificial Intelligence}, vol.~37, no.~2, 2023, pp. 2555--2563.

\bibitem{chen2024topiq}
C.~Chen, J.~Mo, J.~Hou, H.~Wu, L.~Liao, W.~Sun, Q.~Yan, and W.~Lin, ``{TOPIQ}: {A} {Top}-{Down} {Approach} {From} {Semantics} to {Distortions} for {Image} {Quality} {Assessment},'' \emph{IEEE Transactions on Image Processing}, vol.~33, pp. 2404--2418, 2024.

\bibitem{pfiqa}
X.~Lin, X.~Liu, H.~Yang, X.~He, and H.~Chen, ``Perception- and {Fidelity}-{Aware} {Reduced}-{Reference} {Super}-{Resolution} {Image} {Quality} {Assessment},'' \emph{IEEE Transactions on Broadcasting}, vol.~71, no.~1, pp. 323--333, 2025.

\bibitem{zhang2024ldconv}
X.~Zhang, Y.~Song, T.~Song, D.~Yang, Y.~Ye, J.~Zhou, and L.~Zhang, ``{LDConv}: {Linear} deformable convolution for improving convolutional neural networks,'' \emph{Image and Vision Computing}, vol. 149, p. 105190, 2024.

\bibitem{CNNIQA}
L.~Kang, P.~Ye, Y.~Li, and D.~Doermann, ``Convolutional {Neural} {Networks} for {No}-{Reference} {Image} {Quality} {Assessment},'' in \emph{Proceedings of the IEEE/CVF Conference on Computer Vision and Pattern Recognition}, 2014, pp. 1733--1740.

\bibitem{ke2021musiq}
J.~Ke, Q.~Wang, Y.~Wang, P.~Milanfar, and F.~Yang, ``{MUSIQ}: {Multi}-{Scale} {Image} {Quality} {Transformer},'' in \emph{IEEE Proceedings of the IEEE/CVF International Conference on Computer Vision}, 2021, pp. 5148--5157.

\bibitem{tres}
S.~A. Golestaneh, S.~Dadsetan, and K.~M. Kitani, ``No-{Reference} {Image} {Quality} {Assessment} via {Transformers}, {Relative} {Ranking}, and {Self}-{Consistency},'' in \emph{IEEE/CVF Winter Conference on Applications of Computer Vision (WACV)}, 2022, pp. 3989--3999.

\bibitem{livec}
D.~Ghadiyaram and A.~C. Bovik, ``{LIVE} {In} the {Wild} {Image} {Quality} {Challenge} {Database},'' \emph{Online: http://live. ece. utexas. edu/research/ChallengeDB/index. html[Mar, 2017]}, 2015.

\bibitem{hosu2020koniq}
V.~Hosu, H.~Lin, T.~Sziranyi, and D.~Saupe, ``{KonIQ-10k}: {An} {Ecologically} {Valid} {Database} for {Deep} {Learning} of {Blind} {Image} {Quality} {Assessment},'' \emph{IEEE Transactions on Image Processing}, vol.~29, pp. 4041--4056, 2020.

\end{thebibliography}
\end{sloppypar}
\end{document}